\documentclass[10pt,twocolumn,letterpaper]{article}

\usepackage[accsupp]{axessibility}
\usepackage{bm}
\usepackage{color}
\usepackage{multirow}
\usepackage{colortbl}  % for rowcolor
\usepackage{makecell}
\usepackage{pifont}  % for \ding
\usepackage{algorithm}
\usepackage{algorithmic}
\usepackage{listings}
\usepackage[numbers,sort&compress]{natbib}
\newcommand{\myhyperlink}[3][black]{\hyperlink{#2}{\color{#1}{#3}}}

\makeatletter  % as a whole with thickhline
\newcommand{\thickhline}{%
    \noalign {\ifnum 0=`}\fi \hrule height 1pt
    \futurelet \reserved@a \@xhline
}

\definecolor{mygray}{gray}{.9}
\definecolor{ggray}{RGB}{127,127,127}
\definecolor{reda}{RGB}{192,0,0}
\definecolor{redb}{RGB}{217,148,143}
\definecolor{myyellow}{RGB}{190,144,0}
\definecolor{mygreen}{RGB}{80,100,40}
\definecolor{myblue}{RGB}{30,90,100}
\definecolor{mygray1}{RGB}{245,245,245}

\usepackage[pagenumbers]{cvpr} % To force page numbers, e.g. for an arXiv version

\usepackage[dvipsnames]{xcolor}

\definecolor{cvprblue}{rgb}{0.21,0.49,0.74}
\usepackage[pagebackref,breaklinks,colorlinks,citecolor=cvprblue]{hyperref}

\title{Neural Clustering based Visual Representation Learning}

\author{
Guikun Chen\textsuperscript{1}, Xia Li\textsuperscript{2}, Yi Yang\textsuperscript{1}, Wenguan Wang\textsuperscript{1}\footnotemark[1]\\
\small \textsuperscript{1} ReLER, CCAI, Zhejiang University~~\textsuperscript{2} ETH Z\"urich\\
\small\url{https://github.com/guikunchen/FEC/}
}

\begin{document}
\maketitle
\begin{abstract}
\footnotetext[1]{Corresponding author: Wenguan Wang.}
We investigate a fundamental aspect of machine vision: the measurement of features, by revisiting clustering, one of the most classic approaches in machine learning and data analysis. Existing visual feature extractors, including Conv-Nets, ViTs, and MLPs, represent an image as rectangular regions. Though prevalent, such a grid-style paradigm is built upon engineering practice and lacks explicit modeling of data distribution. In this work, we propose \underline{f}eature \underline{e}xtraction with \underline{c}lustering (FEC), a conceptually elegant yet surprisingly ad-hoc interpretable neural clustering framework, which views feature extraction as a process of \textbf{selecting representatives} from data and thus automatically captures the \textbf{underlying data distribution}. Given an ima-ge, FEC alternates between grouping pixels into individual clusters to abstract representatives and updating the deep features of pixels with current representatives. Such an ite-rative working mechanism is implemented in the form of several neural layers and the final representatives can be used for downstream tasks. The cluster assignments across layers, which can be viewed and inspected by humans, make the \textbf{forward} process of FEC \textbf{fully transparent} and empower it with promising ad-hoc interpretability. Extensive experiments on various visual recognition models and tasks veri-fy the effectiveness, generality, and interpretability of FEC. We expect this work will provoke a rethink of the current de facto grid-style paradigm.
\end{abstract}

% \vspace{-8pt}
\section{Introduction}
\label{sec:intro}

\begin{figure}[t]
  \centering
   \includegraphics[width=1.0\linewidth]{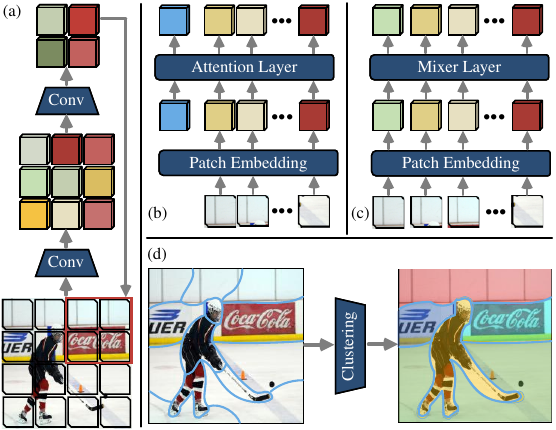}
% \vspace{-15pt}
    \captionsetup{font=small}
  \caption{\small{How to represent an image in a low-dimensional space and what could explain it? (abc) Existing visual backbones rely on the computational modeling of rigid grids. (d) Derived from a neural clustering view, FEC reformulates the procedures of feature extraction as clustering, thereby representing the image with its representatives. Our approach possesses promising \emph{ad-hoc} interpretability and demonstrates the emergence of segmentation despite being trained only on the classification task.}}
  \label{fig:intro}
  % \vspace{-12pt}
\end{figure}

The measurement of features, which explores how to extract abstract, meaningful features from high-dimensional image data, is a topic of enduring interest in machine vision throughout its history~\citep{snyder2004machine,bishop2006pattern,lecun2015deep}. This pursuit, initially dominated by manually engineered descriptors~\citep{ng2003sift,dalal2005histograms,bay2006surf,calonder2010brief,rublee2011orb,leutenegger2011brisk}, has evolved under the influence of deep learning paradigms, transitioning from convolutional landscapes~\citep{Simonyan15,he2016deep} to the frontiers of attention-driven mechanisms~\citep{vaswani2017attention,dosovitskiy2020image} and MLP-based approaches~\citep{tolstikhin2021mlp,touvron2022resmlp}. 
Convolutional networks (Conv-Nets, Fig.~\ref{fig:intro}a) treat an image as rectangular regions and execute in a sliding window manner. Attention-based methods (Fig.~\ref{fig:intro}b) usually divide an image into several non-overlap patches and use an additional [CLS] token to represent the whole image. MLP-based backbones (Fig.~\ref{fig:intro}c) also follow the grid-style paradigm while extracting features without convolution or attention operations.

Upon observing the array of visual backbones shown in Fig.~\ref{fig:intro}, the following questions naturally arise: \hypertarget{Q1}{\ding{182}} \textit{What is the relation between them?} and more critically, \hypertarget{Q2}{\ding{183}} \textit{If these neural networks indeed implicitly capture some intrinsic properties of image data, might there exist a more transparent and interpretable method to measure the visual features?}

The pursuit of the question \myhyperlink{Q1}{\ding{182}} uncovers a persistent adherence to a grid-centric view in the realm of image data analysis~\citep{gonzales1987digital,castleman1996digital,beymer1996image}. Concretely, \textbf{the basic elements during the forward process of existing backbones are rectangular image regions}, \eg, the kernels (filters), sliding window, and receptive field in convolution-based backbones, or the image patches in vision Transformers (ViTs) and MLPs. Such a widespread paradigm, though instrumental in the evolution of convolutional networks and their successors, seems to be based more on engineering convention than on the emulation of natural image structures.
Most existing efforts are expected to generate more abstract features as the network's layers deepen, while nobody knows how they make it~\citep{guidotti2018survey}. Therefore, question \myhyperlink{Q2}{\ding{183}} becomes more fundamental: \hypertarget{Q3}{\ding{184}} \textit{What are the inherent limitations of this grid-style paradigm?} and \hypertarget{Q4}{\ding{185}} \textit{Can we evolve beyond the grid-based uniformity assumption that fails to encapsulate the organic structure of images?}

Driven by question \myhyperlink{Q3}{\ding{184}}, we uncover two critical limitations: \textbf{First}, the grid model is at odds with the true nature of pixel organization, thereby failing to grasp the comple-xity of data distribution~\citep{weinlich2016probability}. \textbf{Second}, the black-box nature of deep feature extractors impedes interpretability, veiling the rationale behind feature selection and significance. This leads us to the question \myhyperlink{Q4}{\ding{185}}, probing the opacity of current methodologies and their divergence from human perception and cognition~\citep{biederman1987recognition,beutter2000motion,bill2020hierarchical}, which possess a remarkable ability to break down visual scenes into semantic-meaning components. The goal is to build feature extractors that can better capture the data distribution of pixels and mirror the cognitive processes of human vision so as to enhance both interpretability and transparency. To bridge the identified gaps, a fundamental paradigm shift is imperative: \textbf{i}) pivo-ting from the grid-view of image representations towards a more fluid model that embraces the dynamic nature of visual data; and \textbf{ii}) stepping away from the black-box models towards an ambitious hybrid that integrates powerful representation learning and interpretable feature encoding. 

In this vein, we introduce FEC (\S\ref{sec:method_fec}), a \emph{mechanistically interpretable} backbone that roots in the principle of clustering. It begins with a window-based pooling to generate pixel blocks that serve as initial elements. Afterward, FEC iterates through two key processes:
\textbf{i}) Clustering-based Feature Pooling. Neural clustering is used to model representatives of the given inputs (pixel blocks or previous clusters), leading to more abstract (\textbf{growing}) clusters. Due to the clustering nature, a representative (cluster) \emph{explicitly} represents a set of pixels in \textbf{any position}. This is where FEC differs from grid-style paradigm. \textbf{ii}) Clustering-based Feature Encoding. Here the representatives are first estimated and then used for redistributing features to each pixel given the similarity between the pixel and its representative. Within such a clustering based framework, the basic elements during FEC's forward process are \textbf{gradually growing clusters}.

FEC exhibits several compelling characteristics: \textbf{First}, enhanced simplicity, and \textbf{\emph{transparency}}. The streamlined design, coupled with the semantic meaning of clustering during feature extraction, renders FEC both conceptually elegant and straightforward to implement. The mechanism by which representatives are modeled ensures that the forward process of FEC is fully transparent. \textbf{Second}, automated discovery of \textbf{\emph{underlying data distribution}}. The deterministic clustering reveals the latent relationships between pixels of the image data, capturing the varying semantic granularity that standard backbones might overlook. As depicted in Fig.~\ref{fig:intro}d, FEC can learn to distinguish non-grid semantic regions autonomously \textit{without explicit supervision of cluster assignments}. \textbf{Third}, \textbf{\emph{ad-hoc interpretability}}. If further inspecting the \emph{cluster assignments} in each feature pooling and combining them together, FEC can interpret its prediction based on the aggregated clusters during forward process and allow users to intuitively view the semantic components. Such \emph{ad-hoc} interpretability is valuable in safety-sensitive scenarios and provides a feasible way for humans to understand the \emph{forward} process of feature extraction. 

By$_{\!}$ answering$_{\!}$ questions$_{\!}$ \myhyperlink{Q1}{\ding{182}}-\myhyperlink{Q4}{\ding{185}},$_{\!}$ we$_{\!}$ formalize$_{\!}$ visual$_{\!}$ feature extraction within a neural clustering-based, fully transpa-rent framework, bridging the gap between classic clustering algorithm and neural network interpretability. We provide a literature review and related discussions in \S\ref{sec:relatedwork}. FEC represents an intuitive and versatile feature extractor, seamlessly compatible with established visual recognition models and tasks, requiring no modifications. Experimental results in \S\ref{sec:exp_cls} show FEC achieves \textbf{72.7}\% \texttt{top-1} accuracy on ImageNet~\citep{ImageNet} with only 5.5M parameters. In \S\ref{sec:exp_ins_rep}, with the modeled representatives, FEC can interpret how it captures the data distribution. In \S\ref{sec:exp_sem_seg} and \S\ref{sec:exp_det}, the transferability and versatility of FEC are validated on three fundamental recognition tasks. Finally, we draw conclusions in \S\ref{sec:conclusion}.

\section{Existing Visual Feature Extractors as Fixed Grid-style Parsers}
\label{sec:backbones}
\noindent\textbf{$_{\!}$Problem$_{\!}$ Statement.$_{\!}$}
Here$_{\!}$ we$_{\!}$ study$_{\!}$ the$_{\!}$ standard$_{\!}$ classifica- tion$_{\!}$ setting.$_{\!}$ Let$_{\!}$ $\mathcal{X}\!$ be$_{\!}$ the$_{\!}$ input$_{\!}$ space$_{\!}$ (\ie,$_{\!}$ image$_{\!}$ space$_{\!}$ for$_{\!}$ vi- sual$_{\!}$ recognition),$_{\!}$ and$_{\!}$ $\mathcal{Y} \!=\! \{\texttt{cat},\cdots,\texttt{dog}\}$ denote the set$_{\!}$ of$_{\!}$ semantic$_{\!}$ categories,$_{\!}$ \eg,$_{\!}$ $|\mathcal{Y}|\!=\! 1000$ for$_{\!}$ ImageNet-1K\citep{ImageNet}.

\noindent\textbf{$_{\!}$Standard$_{\!}$ Pipeline.$_{\!}$} The$_{\!}$ current$_{\!}$ common$_{\!}$ practice$_{\!}$ of$_{\!}$ classifi- cation$_{\!}$ is$_{\!}$ to$_{\!}$ decompose$_{\!}$ the$_{\!}$ deep$_{\!}$ neural$_{\!}$ network$_{\!}$ $h: \mathcal{X} \!\mapsto\! \mathcal{Y}$ into$_{\!}$ $f\!:\! \mathcal{X} \!\mapsto\! \mathcal{F}$ and$_{\!}$ $g\!:\!\mathcal{F} \!\mapsto\! \mathcal{Y}$ that$_{\!}$ $h \!=\! g \circ f$,$_{\!}$ where$_{\!}$ $f\!$ and$_{\!}$ $g$ denote$_{\!}$ the$_{\!}$ feature$_{\!}$ extractor$_{\!}$ and$_{\!}$ the$_{\!}$ classifier,$_{\!}$ respectively.$_{\!}$
Given$_{\!}$ an$_{\!}$ input$_{\!}$ image$_{\!}$ $\bm{X}$, $f$ maps$_{\!}$ it$_{\!}$ into$_{\!}$ a$_{\!}$ \textit{d}-dimensional$_{\!}$ re- presentation$_{\!}$ space$_{\!}$ $\mathcal{F} \!\in\! \mathbb{R}^{C}$, \ie, $\bm{f} \!=\! f(\bm{X})\in\mathbb{R}^{C}$; and $g$ fur- ther predicts the class prediction $\hat{y}$ based on the intermediate feature$_{\!}$ $\bm{f}$,$_{\!}$ \ie,$_{\!}$ $\hat{y}\!=\!g(\bm{f}) \!\in\! \mathcal{Y}$.$_{\!}$ This$_{\!}$ work$_{\!}$ focuses$_{\!}$ on$_{\!}$ the$_{\!}$ $f\!$ only.

\noindent\textbf{ConvNets.$_{\!}$} 
\label{sec:convfe}
Convolution-based$_{\!}$ feature$_{\!}$ extractor$_{\!}$ has$_{\!}$ domina- ted academia and industry for years, whose detailed architectures are reviewed as follows. Formally, given an input image $\bm{X}\! \in\!\mathbb{R}^{3\!\times\! H\!\times\! W}$, ConvNets extract feature embeddings $\{\bm{F}^l\}_{l\!=\!1}^{4}$, where the resolutions are $\frac{1}{4}, \frac{1}{8}, \frac{1}{16}, \frac{1}{32}$ of the original image,$_{\!}$ respectively.$_{\!}$ These$_{\!}$ four$_{\!}$ feature$_{\!}$ embeddings$_{\!}$ are$_{\!}$ generated$_{\!}$ by four separate stages, each containing grid-style feature pooling and encoding.
Taking the 2$^{nd}$ stage of ResNet18~\citep{he2016deep} as an example, given $\bm{F}^1\!\in\!\mathbb{R}^{64\!\times\! 56 \!\times\! 56}$ from the$_{\!}$ 1$^{st}$$_{\!}$ stage,$_{\!}$ a$_{\!}$ low-dimensional$_{\!}$ feature$_{\!}$ map$_{\!}$ is$_{\!}$ generated$_{\!}$ as:$_{\!}$
\vspace{-5pt}
\begin{equation}
\hat{\bm{F}}^2 = \texttt{grid\_pool}(\bm{F}^1)\in\mathbb{R}^{128\!\times\! 28 \!\times\! 28},
\vspace{-5pt}
\end{equation}

\noindent where \texttt{grid\_pool} denotes a convolutional layer with a stride of $2$, which can also be implemented with max poo- ling, average pooling, \etc. After that, feature encoding is performed to get the outputs of this stage:
\vspace{-5pt}
\begin{equation}
\bm{F}^2 = \texttt{encode}(\hat{\bm{F}}^2)\in\mathbb{R}^{128\!\times\! 28 \!\times\! 28},
\vspace{-5pt}
\end{equation}

\noindent where \texttt{encode} denotes several convolutional layers which keep the output resolution consistent. This step is the key essential to distinguish different backbones, which is implemented as self-attention in ViTs and token mixers in MLPs.

\noindent\textbf{ViTs}~\citep{dosovitskiy2020image} and \textbf{MLPs}~\citep{tolstikhin2021mlp}
both commence their operations by generating visual token embeddings for all non-overlap patches of an image:
\vspace{-5pt}
\begin{equation}
    \bm{E} = \texttt{token\_emb}(\bm{X}).
\vspace{-5pt}
\end{equation}

\noindent After which, ViTs use the [CLS] token to represent the whole image while MLPs take the average of all patch embeddings to do so. Since the sequence of image patches is used throughout the forward process of feature extraction, we also categorize them as the grid-style paradigm.

In general, existing backbones are built upon the computational modeling of rigid grids, which uses regular regions to represent an image. However, this paradigm underestimates the dynamic nature of visual scenes, assuming a spatial uniformity that clashes with the \emph{underlying data distribution of pixels}. In addition, it overlooks the essence of human perception, which does not bind itself to rigid grids but instead fluidly navigates through semantic context~\citep{rommers2013context}.

After tackling question \myhyperlink{Q3}{\ding{184}}, in the next section we will detail our clustering based transparent visual feature extractor, which serves as a solid response to question \myhyperlink{Q4}{\ding{185}}.

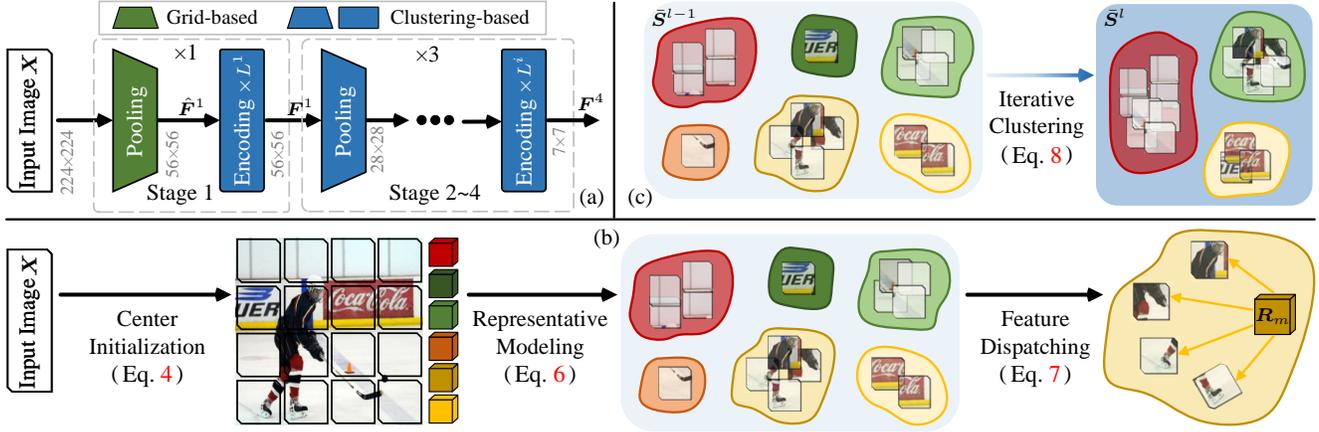
\begin{figure*}[!t]
    \centering
      % \vspace{-10pt}
    \includegraphics[width=1.0\linewidth]{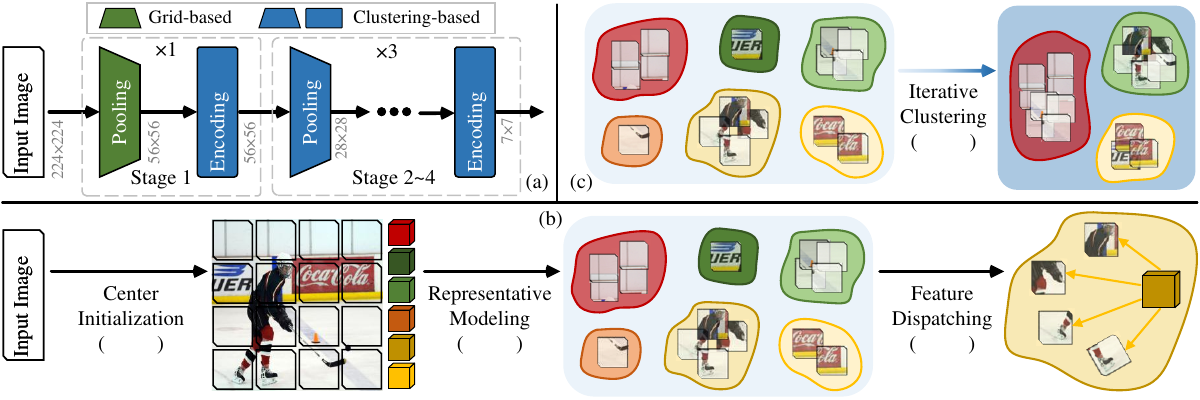}
    \put(-410,129){\scriptsize\rotatebox{90}{\textcolor{white}{$\times L^{1}$}}}
    \put(-304,129){\scriptsize\rotatebox{90}{\textcolor{white}{$\times L^{i}$}}}
    \put(-452,20){\small Eq.~\textcolor{red}{\ref{eq:center_init}}}
    \put(-303.5,20){\small Eq.~\textcolor{red}{\ref{eq:representatives}}}
    \put(-490,60){\footnotesize\rotatebox{90}{\textcolor{black}{$\bm{X}$}}}
    \put(-490,136){\footnotesize\rotatebox{90}{\textcolor{black}{$\bm{X}$}}}
    \put(-23.5,43){\scriptsize\textcolor{black}{$\bm{R}_m$}}
    \put(-431,121){\scriptsize\textcolor{black}{$\hat{\bm{F}}^1$}}
    \put(-390,121){\scriptsize\textcolor{black}{$\bm{F}^1$}}
    \put(-280,123){\scriptsize\textcolor{black}{$\bm{F}^4$}}
    \put(-252,155){\scriptsize\textcolor{black}{$\bar{\bm{S}}^{l-1}$}}
    \put(-81,155){\scriptsize\textcolor{black}{$\bar{\bm{S}}^{l}$}}
    \put(-115.5,103){\small Eq.\textcolor{red}{~\ref{eq:link}}}
    \put(-116,20){\small Eq.~\textcolor{red}{\ref{eq:feat_dis}}}
    
    \vspace{-3pt}
    \captionsetup{font=small}
  \caption{\small{(a) Overall framework of FEC (\S\ref{sec:method_fec}). Each stage $i$ contains $L^{i}$ clustering-based \texttt{encode} layers. (b) Illustration of our clustering-based feature pooling and encoding. (c) The basic elements during FEC's forward process are \emph{growing clusters} instead of \emph{image patches}.}}
    \label{fig:overview}
    \vspace{-8pt}
\end{figure*}

\vspace{-3pt}
\section{Feature Extraction with Clustering (FEC)}
\label{sec:method_fec}

\noindent\textbf{Algorithmic Overview.}
FEC is a neural clustering-based framework for visual feature extraction, building upon the idea of selecting representatives hierarchically. Concretely, given an input image, FEC initiates with a standard convolution whose kernel-size and stride are set to be 4 and 4, respectively. Subsequent feature extraction builds upon these resulted $4\!\times\!4$ pixel patches. Afterward, FEC alternates between the following steps for each given input feature:
\begin{itemize}
    \item Clustering-based Feature \textbf{Encoding}, \ie, $\mathbb{R}^{C\!\times\! W\!\times\!H}\!\mapsto\!\mathbb{R}^{C\!\times\!W\!\times\!H}$. It divides pixels from feature maps into several non-overlap clusters, by projecting the pixel features into a similarity space and using adaptive (the stride and kernel-size are automatically selected to adapt to the desired resolution) average pooling to initialize the cluster centers. As such, cluster assignments can be obtained according to the similarity between pixels and centers. Then, the pixel features are aggregated to construct cluster representations. Subsequently, feature dispatching, which uses the aggregated center to redistribute pixel features within the cluster, is employed to encode pixel-level features, \ie, information communication. Hence elements (pixels) inside the same cluster become more consistent in the feature space.
    \item Clustering-based Feature \textbf{Pooling}, \ie, $\mathbb{R}^{C\!\times\! W\!\times\!H}\!\mapsto\!\mathbb{R}^{C'\!\times\! W/2\!\times\!H/2}$. Similar to the feature encoding process,$_{\!}$ this module uses clustering to obtain the cluster assignments. The difference is that it directly returns the cluster representations to form low-dimensional feature maps without encoding features. These strategies not only preserve the compositional structures of varying semantic$_{\!}$ levels$_{\!}$ but$_{\!}$ also$_{\!}$ seamlessly$_{\!}$ integrate$_{\!}$ the$_{\!}$ concept$_{\!}$ of$_{\!}$ clustering into the feed-forward feature extraction.
\end{itemize}

\noindent To sum up, we formalize our target task --- extracting deep features for visual inputs --- as \textit{representative selection}. By doing so, the intermediate representatives can be a natural substitute for \texttt{grid\_pool}. Since these representatives are computed from the context of each input, it can also be used to communicate information by \textit{feature dispatching}, which serves the same purpose as the \texttt{encode} operation. Next, we will detail the operations of those essential parts of FEC.

\noindent\textbf{$_{\!}$Center$_{\!}$ Initialization.$_{\!}$}
Given$_{\!}$ an$_{\!}$ input$_{\!}$ feature$_{\!}$ map$_{\!}$ $\bm{F} \!\!\in\!\!\mathbb{R}^{N \!\times\! C}$,$_{\!}$ where$_{\!}$ $N\!\!=\!\!W\!\!\times\!\! H$,$_{\!}$ we$_{\!}$ first$_{\!}$ project$_{\!}$ it$_{\!}$ into$_{\!}$ \textit{key}$_{\!}$ and$_{\!}$ \textit{value}$_{\!}$ spaces$_{\!}$ using$_{\!}$ $1 \!\times\! 1\!$ convolutional$_{\!}$ layers,$_{\!}$ resulting$_{\!}$ in$_{\!}$ $\bm{K}\in\mathbb{R}^{N\!\times\!C'\!}$~and$_{\!}$ $\bm{V}\!\in\!\mathbb{R}^{N\!\times\!C'}$, respectively. Here $C'$ is a hyperparameter to control the dimension. We then initialize the cluster centers with their key and value features:
\vspace{-4pt}
\begin{equation}
\label{eq:center_init}
\begin{aligned}
[\bm{C}_{1}^{k}; {\cdots}; \bm{C}_{O}^{k}] &= \texttt{ada\_pool}_{O}(\bm{K})\in\mathbb{R}^{O\!\times\! C'}, \\
[\bm{C}_{1}^{v}; {\cdots}; \bm{C}_{O}^{v}] &= \texttt{ada\_pool}_{O}(\bm{V})\in\mathbb{R}^{O\!\times\! C'},
\end{aligned}
\vspace{-4pt}
\end{equation}
where$_{\!}$ \texttt{ada\_pool}$_{O}\!$ refers$_{\!}$ generating$_{\!}$ $O$$_{\!}$ feature$_{\!}$ centers$_{\!}$ in$_{\!}$ the$_{\!}$ projected$_{\!}$ spaces$_{\!}$ using$_{\!}$ adaptive$_{\!}$ average$_{\!}$ pooling.$_{\!}$ As such,$_{\!}$ the$_{\!}$ centers$_{\!}$ are$_{\!}$ initialized$_{\!}$ adaptively$_{\!}$ for$_{\!}$ each$_{\!}$ input$_{\!}$ itself and gradients can be passed through all indices.

\noindent\textbf{$_{\!}$Representative$_{\!}$ Modeling.$_{\!}$}
To$_{\!}$ \textbf{assign$_{\!}$ each$_{\!}$ element$_{\!}$ into$_{\!}$ a$_{\!}$ cluster},$_{\!}$ the$_{\!}$ similarity$_{\!}$ matrix$_{\!}$ $\bm{M}$ is$_{\!}$ computed$_{\!}$ as:
\vspace{-4pt}
\begin{equation}
\label{eq:compute_sim}
\bm{M} = \langle \bm{K}, [\bm{C}_{1}^{k}; {\cdots}; \bm{C}_{O}^{k}]\rangle \in\mathbb{R}^{N\!\times\! O},
\vspace{-4pt}
\end{equation}

\noindent where $\langle \cdot, \cdot \rangle$ denotes the cosine similarity. Each element is assigned to a cluster \textit{exclusively} according to $\arg\max(\bm{M})$, resulting in an assignment matrix $\bm{A}$ which contains $N$ one-hot$_{\!}$ vectors.$_{\!}$ With$_{\!}$ the$_{\!}$ \textbf{cluster$_{\!}$ assignments},$_{\!}$ the$_{\!}$ deep$_{\!}$ features of the $o$-th representative (cluster) are aggregated by:
\vspace{-4pt}
\begin{equation}
\label{eq:representatives}
\!\!\bm{R}_{o} \!=\! \big(\bm{C}_{o}^{v} \!+ \Sigma_{n=1}^{N}\bm{A}_{no}\bm{V}_{n} \big) /  \big({1+\Sigma_{n=1}^{N}\bm{A}_{no}} \big) \!\in\!\mathbb{R}^{C'}.
\vspace{-4pt}
\end{equation}

\noindent So$_{\!}$ far,$_{\!}$ we've obtained$_{\!}$ the$_{\!}$ low-dimensional$_{\!}$ features$_{\!}$ $\bm{R}\!=\!$ $[\bm{R}_{1}; {\cdots}; \bm{R}_{O}]$ (\ie, representatives), which can seamlessly replace the \texttt{grid\_pool} module in grid-style paradigm.

\noindent\textbf{Feature Dispatching.} \label{sec:feat_dis}
With the insight that elements inside the same cluster shall have similar properties, we propose to propagate the information within each cluster to enhance this phenomenon.
Concretely, we choose to achieve this with modulated propagation with respect to the similarity with the corresponding center~\citep{ma2023image,liang2023clusterfomer}. For element $n$ in cluster $o$, we update its feature $\bm{F}_n\!\in\!\mathbb{R}^{C}$ by:
\vspace{-4pt}
\begin{equation}
\label{eq:feat_dis}
\bm{F}_{n}' = \bm{F}_{n} + \texttt{MLP}(\sigma(\alpha\bm{M}_{no}+\beta)\bm{R}_{o})\in\mathbb{R}^{C},
\vspace{-4pt}
\end{equation}

\noindent where $\sigma$ denotes the sigmoid function. $\alpha$ and $\beta$ are learnable parameters to scale and shift the similarity. The updated features $[\bm{F}_{1}'; {\cdots}; \bm{F}_{N}']$ are the outputs of the \texttt{encode} operation in FEC (can be repeated many times). Since the center features are adaptively sampled from a group of elements, this dispatching process enables effective communication between elements within a cluster and those formed in the cluster center, leading to the overall understanding of the underlying data distribution and context of the image. From a higher perspective, FEC can be seen as an \emph{exclusive} variant (non-overlap clusters) of self-attention~\citep{vaswani2017attention}, \eg, center initialization \emph{vs} key and value matrices, representative modeling \emph{vs} attention scores, and feature dispatching \emph{vs} weighted aggregation. More implementation details are left in the appendix.

\noindent\textbf{Automated Discovery of Underlying Data Distribution.} 
The cluster assignments elucidate not only the relationship between elements and their representatives but also the underlying data distribution within the feature maps. Pixels assigned the $o$-th centroid at the $l$-th level $\{n \mid \bm{A}_{no}^l = 1\}$ coalesce into a cluster (segment) $\bm{S}_o^l$, thereby $\{\bm{S}_o^l \mid 1 \leq o \leq O \}$ decomposing the entire feature map into $O$ discernible segments at the $l$-th layer.
By linking the clusters across sequential layers as follows:
\vspace{-4pt}
\begin{equation}\label{eq:link}
    \bar{\bm{S}}_h^l = \texttt{Union}(\{\bm{S}_o^{l-1} \mid \bm{A}_{oh}^l = 1\}),
\vspace{-4pt}
\end{equation}

\noindent we create a hierarchical pyramid $[\bar{\bm{S}}^1, \bar{\bm{S}}^2, \cdots, \bar{\bm{S}}^L]$
% we ensure connectivity between assignments at various scales. This cross-level linkage gives rise to a hierarchical super-pixels pyramid $[Seg^1, Seg^2, \cdots, Seg^L]$, 
which coalesces pixels into increasingly larger segments, and explicitly reveals the underlying data distribution.

\noindent\textbf{\emph{Ad-hoc} Interpretability.} 
The described configuration facilitates a direct forward process $l\!\!=\!\!1\!\!\to\!\!L$ that yields a linked spatial decomposition $[\bar{\bm{S}}^1, \bar{\bm{S}}^2, \cdots, \bar{\bm{S}}^L]$, intuitively conveying image parsing to the observer. In contrast, earlier techniques (\eg, Grad-CAM~\citep{selvaraju2017grad}) demand a retrospective process to accentuate the activated regions. These methods typically require complex post-processing to elucidate the concealed parsing mechanism. However, FEC is \emph{mechanistically interpretable} because its forward process based on gradually growing clusters (segments) is \emph{fully transparent}. See~\citep{rudin2022interpretable} for a more detailed discussion.

\noindent\textbf{Versatility.} 
After modeling the feature extraction process with representatives, one may wonder about the applicability of this new paradigm to dense prediction tasks. For instance, in the detection task, the training process of widely-used models like YOLO~\citep{redmon2016you} and Faster-RCNN~\citep{ren2015faster}, relies on the grid-based label assignments (anchors). To keep the necessary grid-based information, we introduce the residual connection~\citep{he2016deep} to the feature of representatives:
\vspace{-4pt}
\begin{equation}
\label{eq:residual}
\bm{R}_{i} = \texttt{ResConn}(\bm{F}_{i}) + \bm{R}_{i} \in\mathbb{R}^{C'}.
\vspace{-4pt}
\end{equation}

\noindent This modification allows each $\bm{R}_{i}\!\in\!\mathbb{R}^{C'}$ to represent both the rectangular regions in grid-style paradigms and the selected representatives in our clustering-based paradigm. As shown in Fig.~\ref{fig:overview}a, we also adopt four stages during feature extraction as in existing standard backbones, resulting in the same resolutions of output features. In a nutshell, FEC marks a fundamental paradigm shift for visual feature extraction while retaining full compatibility with previous works. The computational pipeline is also detailed in \S\ref{sec:exp_ins_rep}. 

\noindent\textbf{Adaptation to Downstream Tasks.} 
As aforementioned, with the introduction of residual connections, FEC can be seamlessly incorporated into dense prediction tasks like detection and segmentation without any architectural modifications. In terms of classification, we use the standard classification head over the final feature map $\bm{F}^L$, \ie, a single-layer MLP, which takes the average of all representatives. More$_{\!}$ details$_{\!}$ are$_{\!}$ left$_{\!}$ in$_{\!}$ the$_{\!}$ appendix.$_{\!}$ We$_{\!}$ expect$_{\!}$ the$_{\!}$ recent$_{\!}$ set-prediction architectures (\eg, DETR~\citep{carion2020end}) can better utilize the modeled representatives and leave it as future works.

\section{Related Work}
\label{sec:relatedwork}
In this section, we review representative work in clustering, visual$_{\!}$ feature$_{\!}$ extraction,$_{\!}$ and$_{\!}$ neural$_{\!}$ network$_{\!}$ interpretability.
 
\noindent\textbf{Clustering}
stands as a foundational technique in machine learning that involves grouping akin data points together based on their intrinsic characteristics. Given numerous data, the goal of clustering is to model meaningful clusters (given pre-defined numbers or not), which can be viewed as the summary of the raw data. Subsequently, the similarities between each data and the cluster representations can be quantified. Clustering has been widely used in various fields, \eg, scene understanding~\citep{li2022nicest,li2022devil,chen2023addressing,li2023compositional}, point clouds~\citep{yin2022proposalcontrast,feng2023clustering,feng2024interpretable3D}, segmentation~\citep{wang2018semi,liang2022gmmseg,zhou2022rethinking,liang2023clustseg,li2023unified}, and AI4Science~\citep{grabski2023significance,barrio2023clustering,wayment2024predicting,quan2024clustering}.

In a departure from previous endeavors that harnessed clustering mechanisms as add-on heads to facilitate specific tasks, our approach pioneers the idea of learning universal visual representations from the clustering view. The proposed clustering-based feature extraction shares a kinship with the classic vision technique of clustering similar pixels~\citep{ren2003learning}. Nonetheless, our innovation lies in its capacity to capture underlying data distribution and generate continuous representations for downstream tasks. Via reformulating the entire process of feature extraction as selecting representatives in a clustering fashion, FEC capitalizes on the robustness of end-to-end representation learning while preserving the transparent nature inherent to clustering. Inspired by biological systems that process inputs from different modalities simultaneously, Perceivers~\citep{jaegle2021perceiver,carreira2022hip} model a set of \emph{latent vectors} correlated with inputs. FEC is motivated from the classic idea of clustering, and its intermediate elements have \emph{explicit} meaning, \ie, segments (Eq.~\ref{eq:link}).

\noindent\textbf{$_{\!}$Feature Extractors for Vision.}$_{\!}$ 
Feature extraction is a pivotal aspect of machine learning, serving as a crucial step in transforming raw data into informative representations conducive to computational analysis. While some pre-deep learning methods~\citep{gu2009recognition,carreira2012semantic} also aim to get rid of grids, they are not \emph{fully} end-to-end and do not scale well. 
Since the 2010s, convolutional networks ~\citep{krizhevsky2012imagenet} have reigned supreme in computer vision, with significant milestones like VGG~\citep{Simonyan15}, ResNet~\citep{he2016deep}, ConvNeXt~\citep{liu2022convnet} and so on~\citep{szegedy2015going,huang2017densely,tan2019efficientnet} consistently pushing the envelope by deepening and optimizing convolutional layers, enabling the extraction of hierarchical and abstract features.$_{\!}$ Innovations$_{\!}$ like$_{\!}$ depthwise$_{\!}$ convolution$_{\!}$~\citep{howard2017mobilenets}$_{\!}$ and$_{\!}$ deformable convolution~\citep{dai2017deformable} have further enhanced feature extraction within$_{\!}$ these$_{\!}$ networks.$_{\!}$
More recently,$_{\!}$ the breakthroughs in NLP ushered in a revolution in feature extraction, introducing$_{\!}$ attention-based$_{\!}$ Transformer$_{\!}$ architectures$_{\!}$~\citep{vaswani2017attention}.$_{\!}$ Vision$_{\!}$ Transformer$_{\!}$ (ViT)$_{\!}$ is$_{\!}$ a$_{\!}$ time-honored$_{\!}$ work$_{\!}$~\citep{dosovitskiy2020image},$_{\!}$ which$_{\!}$ adopts$_{\!}$ self-attention$_{\!}$ mechanisms$_{\!}$ to$_{\!}$ image$_{\!}$ classification,$_{\!}$ achieving$_{\!}$ remarkable results.$_{\!}$ Note$_{\!}$ that$_{\!}$ ViTs'$_{\!}$ effectiveness$_{\!}$ often$_{\!}$ hinges$_{\!}$ on large-scale training datasets~\citep{khan2022transformers,gani2022train}. Recent advancements have endeavored to bridge the gap between convolution and attention through hybrid models like CoAtNet~\citep{dai2021coatnet} and Mobile-Former~\citep{chen2022mobile}, which seamlessly combine both design paradigms. Besides, models employing multi-layer perceptrons (MLPs)~\citep{tolstikhin2021mlp,touvron2022resmlp,guo2022hire,hou2022vision,tang2022image,chen2021cyclemlp} for spatial interactions and techniques such as shifting~\citep{huang2021shuffle,lian2021mlp} or pooling~\citep{yu2022metaformer} for local context $_{\!}$ have$_{\!}$ emerged$_{\!}$ to$_{\!}$ explore$_{\!}$ the$_{\!}$ power$_{\!}$ of$_{\!}$ MLPs.

Though impressive, existing feature extractors are built upon rectangular receptive fields (as in CNN and variants, which represent an image as a grid of regular regions), or queries (as in ViTs, which incorporate an additional query token to represent an image), or rectangular image patches (as in MLPs, which use mixer layers to perform communication between patches). As far as we know, CoC~\citep{ma2023image} is the only backbone using a clustering algorithm. Compared to CoC, we want to highlight that our essential paradigm (\ie, representing an image as representatives \emph{vs} rectangular regions), goal (\ie, \emph{transparent} and \emph{ad-hoc interpretable} backbone \emph{vs} clustering based method for information exchange), and core techniques (\ie, modeling \emph{gradually growing} representatives \emph{vs} modeling context clusters at specific resolution) are different. FEC takes the \emph{first} step towards \emph{fully} clustering based feature extraction, by reformulating its workflow as selecting representatives.

\noindent\textbf{Neural Network Interpretability.} 
The opaque nature of deep neural networks (DNNs) has posed challenges to their adoption in decision-critical applications, sparking a surge of interest in enhancing their transparency and interpretabi-lity. One key distinction in interpretable methods lies in whether they produce posterior explanations for pre-trained DNNs$_{\!}$ or$_{\!}$ aim$_{\!}$ to$_{\!}$ develop$_{\!}$ inherently$_{\!}$ interpretable$_{\!}$ DNNs$_{\!}$ from$_{\!}$ the$_{\!}$ outset.$_{\!}$
Posterior$_{\!}$ explanations,$_{\!}$ though$_{\!}$ prevalent,$_{\!}$ face$_{\!}$ cri- ticism for their nature of approximation and limited capaci- ty to truly elucidate the inner workings of DNNs~\citep{laugel2019dangers,rudin2019stop,arrieta2020explainable,rudin2022interpretable,chen2019looks}. Representative works involve reverse-engineer importance values~\citep{erhan2009visualizing,simonyan2013deep,zeiler2014visualizing,selvaraju2017grad,mahendran2015understanding,yosinski2015understanding,bach2015pixel,cam2016learning,shrikumar2017learning} and$_{\!}$ sensitivities$_{\!}$ of inputs~\citep{ribeiro2016should,zintgraf2017visualizing,koh2017understanding,lundberg2017unified}. Recent advances in self-supervised learning, such as DINO~\citep{caron2021emerging,oquab2023dinov2}, have revealed$_{\!}$ emergent$_{\!}$ segmentation$_{\!}$ properties$_{\!}$ in$_{\!}$ ViTs.$_{\!}$ On$_{\!}$ the$_{\!}$ other$_{\!}$ hand,$_{\!}$ methods$_{\!}$ focused$_{\!}$ on$_{\!}$ ad-hoc$_{\!}$ explainability$_{\!}$ strive$_{\!}$ to$_{\!}$ incorporate$_{\!}$ more$_{\!}$ interpretable elements into black-box DNNs~\citep{alvarez2018towards,kim2018interpretability,wang2019gaining} or impose specific properties on the model's representations~\citep{subramanian2018spine,chen2016infogan,you2017deep}$_{\!}$ to$_{\!}$ bolster$_{\!}$ interpretability.$_{\!}$ A$_{\!}$ notable$_{\!}$ instance$_{\!}$ is the deep nearest centroids~\citep{wang2023visual}, which introduces a nonparametric classifier and uses case-based reasoning, thereby presenting a novel and explainable paradigm for visual recognition.

In$_{\!}$ a$_{\!}$ related$_{\!}$ vein,$_{\!}$ CRATE$_{\!}$~\citep{yu2023emergence}$_{\!}$ adopts$_{\!}$ the$_{\!}$ white-box Transformer$_{\!}$~\citep{yu2023white} to$_{\!}$ delve$_{\!}$ into$_{\!}$ visual$_{\!}$ attention$_{\!}$ and$_{\!}$ the$_{\!}$ emer- gence of segmentation on the supervised classification task, showcasing$_{\!}$ considerable results.$_{\!}$ Nonetheless,$_{\!}$ we argue that CRATE$_{\!}$ falls$_{\!}$ within$_{\!}$ the$_{\!}$ realm$_{\!}$ of$_{\!}$ posterior$_{\!}$ explanations,$_{\!}$ whereas our FEC possesses an inherent capacity to elucidate the reasons behind its feature extraction process. Such innate transparency sets FEC apart from counterparts that solely offer \emph{post-hoc} explainability. This work represents a small yet solid stride towards empowering the forward process of feature extraction with \emph{ad-hoc} interpretability through an integrated clustering-based backbone and offers a fresh perspective on deep visual representations.

\section{Experiment}

FEC is proposed as the first framework to support feature extraction in a fully clustering-based manner. We evaluate the proposed backbone for four recognition tasks \textit{viz}. image classification, object detection, semantic segmentation, and instance segmentation on three prevalent benchmarks (\ie, ImageNet-1K~\citep{ImageNet}, MS COCO~\citep{lin2014microsoft}, and ADE20k~\citep{zhou2017scene}).

FEC combines \emph{ad-hoc} interpretability with promising performance$_{\!}$ across$_{\!}$ diverse$_{\!}$ recognition$_{\!}$ tasks$_{\!}$ and$_{\!}$ datasets.$_{\!}$ Note that our objective is not to chase the state-of-the-art$_{\!}$ performance$_{\!}$ like$_{\!}$ ConvNeXt~\citep{liu2022convnet},$_{\!}$ but$_{\!}$ rather$_{\!}$ to$_{\!}$ assess$_{\!}$ FEC's$_{\!}$ capabilities$_{\!}$ (\eg,$_{\!}$ effectiveness,$_{\!}$ efficiency,$_{\!}$ transferability,$_{\!}$ \etc.) through a comprehensive series of experiments.

\subsection{Experiments on Image Classification}
\label{sec:exp_cls}

\noindent\textbf{Dataset.$_{\!}$}
ImageNet-1K~\citep{ImageNet} is a well-benchmarked image dataset. Following conventional procedures, it is divided into 1.28M/50K/100K images for \texttt{train}/\texttt{val}/\texttt{test}.

\noindent\textbf{Training.}
We use \textit{timm}~\citep{rw2019timm} as our codebase and follow the standard training protocols as detailed in~\citep{dai2021coatnet,yu2022metaformer,ma2023image}. We use an AdamW~\citep{loshchilov2017decoupled} optimizer using a cosine decay learning rate scheduler and 5 epochs of warm-up. The momentum and weight decay are set to 0.9 and 0.05, respectively. A batch size of 1024 and an initial learning rate of 0.001 are used. More details are left in the appendix.

\noindent\textbf{Test.} Following~\citep{ma2023image}, we use one input image scale of 224$\times$224 with center cropping without any data augmentation. By default, the models are trained on 4 NVIDIA Tesla A100 GPUs with 80GB memory per card. Testing is conducted on the same machine.

\noindent\textbf{Metric.}
We report the parameters used, FLOPs, and Top-1 classification accuracy on a single crop following previous works~\citep{he2016deep,liu2021swin}. Throughput (image/s), or FPS, is measured using the same script as in~\citep{liu2021swin,touvron2021training} on a single V100 GPU using a batch size of 256.

\begin{table}[!t]
    \centering
    \small
    \caption{Quantitative results on ImageNet-1K~\citep{ImageNet} \texttt{val} for \textbf{ima-ge classification} (\S\ref{sec:exp_cls}). All models are trained and tested at 224$\times$224 resolution, except ViT-B~\citep{dosovitskiy2020image} and ViT-L~\citep{dosovitskiy2020image}. }
    \scalebox{0.94}{
    \setlength\tabcolsep{2pt}
    \label{tab:imagenet_cls}
    \begin{tabular}{|cl||cccc|}
        \hline\thickhline
        \rowcolor{mygray} & Method & \makecell{ \#Param\\(M)} & \makecell{ FLOPs\\(G)} & \makecell{ \quad Top-1\quad\quad\\(\%)$\uparrow$} & \makecell{ FPS\\(img/s)$\uparrow$}  \\
        \hline\hline
        \multirow{3}{*}{\rotatebox{90}{\textbf{Clu.}}} & CoC-Tiny$_{\!\!}$~~~~~~~~~~~~~~~\citep{ma2023image}  & 5.3 &1.1 &71.8 &1146.7\\
         & CoC-Small$_{\!\!}$~~~~~~~~~~~~~\citep{ma2023image} & 14.0 &2.8 &77.5 &852.1\\
         & CoC-Medium~~~~~~~~\citep{ma2023image} & 27.9 &5.9 &81.0  &345.7\\
        \hline\hline
         \multirow{7}{*}{\rotatebox{90}{\textbf{MLP}}}& ResMLP-12$_{\!}$~~~~~~~~~~~\citep{touvron2022resmlp} & 15.0 & 3.0 & 76.6  &1499.0 \\
         & ResMLP-24$_{\!}$~~~~~~~~~~~\citep{touvron2022resmlp} & 30.0 & 6.0 & 79.4  &741.6\\
         & ResMLP-36$_{\!}$~~~~~~~~~~~\citep{touvron2022resmlp} & 45.0 & 8.9 & 79.7  &484.6 \\
         & MLP-Mixer-B/16~~\citep{tolstikhin2021mlp} & 59.0 & 12.7 & 76.4  &387.7 \\
         & MLP-Mixer-L/16~~\citep{tolstikhin2021mlp} & 207.0 & 44.8 & 71.8  &114.2 \\
         & gMLP-Ti~~~~~~~~~~~~~~~\citep{liu2021pay} & 6.0 & 1.4 & 72.3  &1440.0 \\
         & gMLP-S~~~~~~~~~~~~~~~~\citep{liu2021pay} & 20.0 & 4.5 & 79.6  &650.5 \\
        \hline\hline
         \multirow{6}{*}{\rotatebox{90}{\textbf{Attention}}}& ViT-B/16~~~~~~~~~~~~~~~\citep{dosovitskiy2020image} &86.0  &55.5 &77.9  & 86.4\\
         & ViT-L/16~~~~~~~~~~~~~~~\citep{dosovitskiy2020image} & 307 &190.7 &76.5  & 26.6\\
         & PVT-Tiny~~~~~~~~~~~~~~\citep{wang2021pyramid} & 13.2 & 1.9 &75.1 &- \\
         & PVT-Small~~~~~~~~~~~~\citep{wang2021pyramid}  & 24.5 & 3.8 & 79.8  &-\\
        & Swin-Tiny~~~~~~~~~~~~~\citep{liu2021swin}  & 29 & 4.5 & 81.3 & 631.1\\
        & Swin-Small~~~~~~~~~~~\citep{liu2021swin}  & 50 & 8.7 & 83.0 & 374.9\\
        \hline\hline
         \multirow{5}{*}{\rotatebox{90}{\textbf{Conv.}}}& ResNet18~~~~~~~~~~~~~~\citep{he2016deep} & 12 & 1.8 & 69.8 &  4284.9  \\
         & ResNet50~~~~~~~~~~~~~~\citep{he2016deep} & 26 & 4.1 & 79.8  & 1206.0  \\
         & ConvMixer$_{512/16\!}$~~\citep{trockman2022patches} & 5.4 & - & 73.8 & -  \\
         & ConvMixer$_{1024/12}$\citep{trockman2022patches} & 14.6 & - & 77.8 & -  \\
         & ConvMixer$_{768/32\!}$~~\citep{trockman2022patches} & 21.1 & - & 80.2 & 139.7  \\
        \hline\hline 
        \multirow{3}{*}{\rotatebox{90}{\textbf{Clu.}}} & FEC-Small (Ours)  & 5.5 &1.4 & 72.7$_{\textcolor{gray}{\pm0.06}}$ &1042.9\\
         & FEC-Base ~~$_{\!}$(Ours) & 14.4 &3.4 & 78.1$_{\textcolor{gray}{\pm0.00}}$ &754.1\\
         & FEC-Large (Ours) & 28.3 &6.5 & 81.2$_{\textcolor{gray}{\pm0.06}}$  &342.1\\
        \hline
    \end{tabular}
    }
\end{table}

\noindent\textbf{Performance$_{\!}$ Comparison.$_{\!}$}
Table$_{\!}$~\ref{tab:imagenet_cls}$_{\!}$ presents$_{\!}$ our$_{\!}$ classifi- cation$_{\!}$ results$_{\!}$ on$_{\!}$ ImageNet$_{\!}$~\citep{ImageNet}$_{\!}$ \texttt{val}.$_{\!}$ As illustrated, FEC achieves competitive performance and efficiency, outperforming ResNet18~\citep{he2016deep} by \textbf{2.9}\% using less than half the number of parameters that ResNet18 has. ResNet achieves high FPS due to optimizations for convolutional operators.

\begin{figure*}[t]
    \centering
   \includegraphics[width=1.0\linewidth]{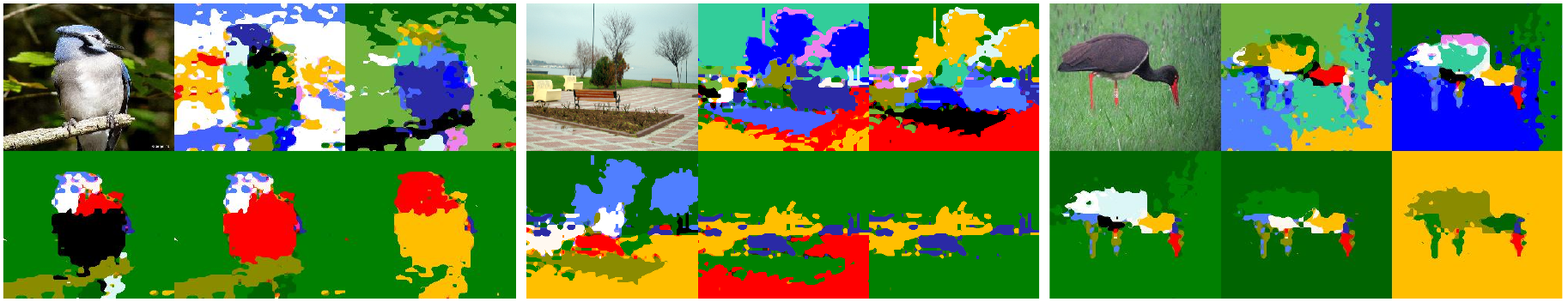}
     \vspace{-15pt}
   \caption{Inspection of the modeled representatives (\S\ref{sec:exp_ins_rep}) on ImageNet-1K~\citep{ImageNet} \texttt{val}. Different colored masks indicate different clusters. As the number of clusters decreases, each cluster tends to represent an entire object or a portion of an object, suggesting that FEC effectively captures the underlying data distribution of visual scenes. }
  \vspace{-1pt}
\label{fig:ins_rep}
\end{figure*}

\begin{figure*}
\vspace{-3pt}

\begin{minipage}[l]{0.3\linewidth}
\captionsetup{type=table}
\centering
\small
\caption{$_{\!}$Quantitative$_{\!}$ results$_{\!}$ on$_{\!}$ ADE20K \citep{zhou2017scene}$_{\!}$ \texttt{val}$_{\!}$ for$_{\!}$ \textbf{semantic segmentation} (\S\ref{sec:exp_sem_seg}). Semantic$_{\!}$ FPN~\citep{kirillov2019panoptic} is adopted.}
\vspace{-3pt}
\label{tab:ade20k}
\scalebox{0.83}{
\setlength\tabcolsep{4pt}
\begin{tabular}{|l||c|c|}
    \hline\thickhline
    \rowcolor{mygray} Backbone & \makecell{ \#Param\\(M)} &   \makecell{ ADE20K \\mIoU(\%)$\uparrow$} \\
    \hline\hline
    ResNet18~~~~~~~~~~~~\citep{he2016deep} & 15.5M & 32.9 \\
    ResNet50~~~~~~~~~~~~\citep{he2016deep} & 28.5M & 36.7 \\
    PVT-Tiny~~~~~~~~~~~~\citep{wang2021pyramid} & 17.0M & 35.7\\
    PVT-Small~~~~~~~~~~\citep{wang2021pyramid} & 28.2M & 39.8\\
    CoC-Small/4~~~~~~~\citep{ma2023image} & 17.6M & 36.6 \\
    CoC-Small/25~~~~~\citep{ma2023image} & 17.6M & 36.4 \\
    CoC-Small/49~~~~~\citep{ma2023image} & 17.6M & 36.3 \\
    CoC-Medium/4~~~\citep{ma2023image} & 31.5M & 40.2 \\
    CoC-Medium/25~\citep{ma2023image} & 31.5M & 40.6 \\
    CoC-Medium/49~\citep{ma2023image} & 31.5M & 40.8 \\
    \hline\hline
    FEC-Small & 9.1M & 35.3$_{\textcolor{gray}{\pm0.15}}$ \\
    FEC-Base & 18.0M & 37.7$_{\textcolor{gray}{\pm0.15}}$ \\
    FEC-Large & 31.9M & 40.5$_{\textcolor{gray}{\pm0.10}}$ \\
    \hline
\end{tabular}}
\end{minipage}\hspace{7pt}
\begin{minipage}[r]{0.675\linewidth}
\centering
\small
\captionsetup{type=table}
\caption{Quantitative results on COCO~\citep{lin2014microsoft} \texttt{val2017} for \textbf{object detection} and \textbf{semantic segmentation} (\S\ref{sec:exp_det}). We use Mask RCNN~\citep{he2017mask} to evaluate the performance of the proposed backbone on two tasks.}
\vspace{-3pt}
\label{tab:det_seg}
\scalebox{0.83}{
\setlength\tabcolsep{3.5pt}
\begin{tabular}{|l||c|ccc|ccc|}
    \hline\thickhline
    \rowcolor{mygray} Backbone& \makecell{ \#Param\\(M)}  & \makecell{  \\ ~AP$^{\mathrm{box}}\uparrow$}~ & \makecell{  COCO\\ ~AP$^{\mathrm{box}}_{50}\uparrow$}~  & \makecell{  \\ ~AP$^{\mathrm{box}}_{75}\uparrow$}~ & \makecell{  \\ ~AP$^{\mathrm{mask}}\uparrow$}~ & \makecell{  COCO\\ ~AP$^{\mathrm{mask}}_{50}\uparrow$}~ & \makecell{  \\ ~AP$^{\mathrm{mask}}_{75}\uparrow$}~ \\
    \hline\hline
    ResNet18~~~~~~~~~~~~\citep{he2016deep} & 31.2M &34.0  &54.0 & 36.7 &31.2  &51.0 &32.7\\
    ResNet50~~~~~~~~~~~~\citep{he2016deep} & 44.2M & 38.0 & 58.6 & 41.4 & 34.4 & 55.1 & 36.7\\
    PVT-Tiny~~~~~~~~~~~~\citep{wang2021pyramid} & 32.9M &36.7  &59.2 &39.3 &35.1  &56.7 &37.3\\
    PVT-Small~~~~~~~~~~\citep{wang2021pyramid} & 44.1M & 40.4 & 62.9 & 43.8 & 37.8 & 60.1 & 40.3\\
    CoC-Small/4~~~~~~~\citep{ma2023image} & 32.7M &35.9 & 58.3 & 38.3 & 33.8 & 55.3 & 35.8\\
    CoC-Small/25~~~~~\citep{ma2023image} & 32.7M &37.5 & 60.1 & 40.0 & 35.4 & 57.1 & 37.9\\
    CoC-Small/49~~~~~\citep{ma2023image} & 32.7M &37.2 & 59.8 & 39.7 & 34.9 & 56.7 & 37.0\\
    CoC-Medium/4~~~\citep{ma2023image} & 46.7M &38.6 & 61.1 & 41.5 & 36.1 & 58.2 & 38.0\\
    CoC-Medium/25~\citep{ma2023image} & 46.7M &40.1 & 62.8 & 43.6 & 37.4 & 59.9 & 40.0\\
    CoC-Medium/49~\citep{ma2023image} & 46.7M &40.6 & 63.3 & 43.9 & 37.6 & 60.1 & 39.9\\
    \hline\hline
    FEC-Small & 24.3M & 35.6$_{\textcolor{gray}{\pm0.06}}$ & 57.5$_{\textcolor{gray}{\pm0.10}}$ & 38.2$_{\textcolor{gray}{\pm0.15}}$ & 33.6$_{\textcolor{gray}{\pm0.10}}$  & 54.7$_{\textcolor{gray}{\pm0.10}}$ & 35.7$_{\textcolor{gray}{\pm0.15}}$\\
    FEC-Base & 33.1M & 37.9$_{\textcolor{gray}{\pm0.06}}$ & 60.1$_{\textcolor{gray}{\pm0.06}}$ & 40.8$_{\textcolor{gray}{\pm0.15}}$ & 35.5$_{\textcolor{gray}{\pm0.06}}$  & 57.2$_{\textcolor{gray}{\pm0.06}}$ & 37.8$_{\textcolor{gray}{\pm0.10}}$\\
    FEC-Large & 47.1M & 39.9$_{\textcolor{gray}{\pm0.10}}$ & 62.5$_{\textcolor{gray}{\pm0.06}}$ & 43.2$_{\textcolor{gray}{\pm0.12}}$ & 37.3$_{\textcolor{gray}{\pm0.06}}$  & 59.5$_{\textcolor{gray}{\pm0.15}}$ & 39.5$_{\textcolor{gray}{\pm0.12}}$\\
    \hline
\end{tabular}}
\end{minipage}

\vspace{-8pt}
\end{figure*}

\subsection{Study of {Ad-hoc} Interpretability}
\label{sec:exp_ins_rep}
We have empirically demonstrated FEC's effectiveness and efficiency in image classification. By redefining visual feature extraction as a clustering process, where representatives are iteratively selected during the forward process, FEC departs from the traditional grid-style paradigms. This shift suggests that FEC offers exceptional \emph{ad-hoc} interpretability, a quality we will now explore further.

We first detail the procedure to aggregate cluster assignments across layers. Given an image (224$\times$224), a standard convolution-based pooling is used to generate a low-dimensional feature map (56$\times$56) where each pixel represents a 4$\times$4 region in the raw image. 
As for the first clustering-based pooling layer, those \textbf{pixel blocks} (56$\times$56, each with 4$\times$4 pixels) will be assigned to one of a total of 28$\times$28 clusters $\{\bm{S}_i^1,\!1\!\leq \!i\!\leq\!28\!\times\!28\}$. Afterward, the subsequent pooling layers perform the same clustering procedure based on \textbf{previous clusters}. In this way, the clusters at different levels can be aggregated across layers according to Eq.~\ref{eq:link}, leading to 7$\times$7 segments $\{\bar{\bm{S}}_i^4,\!1\!\leq \!i\!\leq\!7\!\times\!7\}$ in the last step. While humans can view and examine the final cluster assignments, 49 clusters are simply too many to comprehend for a 224$\times$224 image. Therefore, we use K-Means to further reduce the number of clusters. Concretely, we use the deep features ($\bm{R}$ in Eq.~\ref{eq:representatives}) of those 49 clusters as their representations. Default hyperparameters in the scikit-learn \citep{scikit-learn} implementation are adopted. Medium filtering is adopted for better visualization.

After the above introduction, we visualize cluster assignments in Fig.~\ref{fig:ins_rep} to clarify FEC's principles. Two remarkable observations justify the \emph{ad-hoc} interpretability and effectiveness of FEC: \textbf{i}) Across successive pooling layers, the clustering-based feature extraction progressively abstracts pixel blocks, mirroring human cognitive processes and providing a natural representation of visual data.  \textbf{ii}) In final cluster assignments, we find consistent semantic representations, with clusters often corresponding to coherent objects or object parts. This aligns FEC with human perception and enhances feature interpretability, linking clusters to identifiable image elements.

\begin{table*}[t]
\caption{A  set  of  ablative  experiments (\S\ref{sec:exp_cls})  on  ImageNet~\citep{ImageNet} \texttt{val}, ADE20K~\citep{zhou2017scene} \texttt{val}, and COCO~\citep{lin2014microsoft} \texttt{val2017}. The adopted hyperparameters are marked in \textcolor{red}{red}.}
\vspace{-4pt}

\begin{subtable}{0.31\linewidth}
\scalebox{0.85}{
\setlength\tabcolsep{4pt}
\begin{tabular}{|c||c|cc|}
    \hline\thickhline
    \rowcolor{mygray} \makecell{ similarity \\ measurement}& \makecell{ \makecell{ \#Param.\\(M)}} & \makecell{ Top-1\\(\%)$\uparrow$} & \makecell{ Top-5\\(\%)$\uparrow$} \\
    \hline\hline
    Euclidean & 5.46 & 72.2 & 90.9 \\
    Dot Product & 5.46 & 72.7 & 91.1 \\
    \textcolor{red}{Cosine} & 5.46 & 72.7 & 91.2 \\
    \hline
\end{tabular}
}
\vspace{5pt}
\centering
\small
\caption{Similarity measurement (Eq.~\ref{eq:compute_sim})}
\label{tab:ab_sim}
\end{subtable}\hspace{1pt}
\begin{subtable}{0.32\linewidth}
\scalebox{0.85}{
\setlength\tabcolsep{4pt}
\begin{tabular}{|c||c|cc|}
    \hline\thickhline
    \rowcolor{mygray} \makecell{ ~~~~~feature~~~~ \\ ~~~~~dispatching~~~~}& \makecell{ \makecell{ \#Param.\\(M)}} & \makecell{ Top-1\\(\%)$\uparrow$} & \makecell{ Top-5\\(\%)$\uparrow$} \\
    \hline\hline
    \textit{w/o} $\alpha$ and $\beta$ & 5.46 & 72.1 & 90.8 \\
    \textit{w/o} $\alpha$ & 5.46 & 72.4 & 91.0 \\
    \textit{w/o} $\beta$ & 5.46 & 72.2 & 90.9 \\
    \hline
    Ours (Eq.~\ref{eq:feat_dis}) & 5.46 & 72.7 & 91.2 \\
    \hline
\end{tabular}
}
\vspace{0.2pt}
\centering
\small
\caption{Parameters used in feature dispatching (\S\ref{sec:feat_dis})}
\label{tab:ab_feat_dis}
\end{subtable}\hspace{6pt}
\begin{subtable}{0.32\linewidth}
\scalebox{0.85}{
\setlength\tabcolsep{4pt}
\begin{tabular}{|c||c|c|}
    \hline\thickhline
    \rowcolor{mygray} feature & \#Param & ADE20K  \\
    \rowcolor{mygray} dimension & (M) & mIoU(\%)$\uparrow$ \\
    \hline\hline
    (48, 48, 96, 96) & 8.6M & 34.9 \\
    (72, 72, 144, 144) & 8.9M & 35.2 \\
    \textcolor{red}{(96, 96, 192, 192)} & 9.1M & 35.3 \\
    (120, 120, 240, 240) & 9.4M & 35.5 \\
    \hline
\end{tabular}}
\centering
\small
\vspace{0.2pt}
\caption{Number of channels in \texttt{encode} (Eq.~\ref{eq:center_init})}
\label{tab:ab_seg}
\end{subtable}

\centering
\begin{subtable}{0.98\linewidth}
\scalebox{0.85}{
\setlength\tabcolsep{4pt}
\begin{tabular}{|c||c|cccccc|ccc|}
    \hline\thickhline
    \rowcolor{mygray} feature & \#Param  & \multicolumn{6}{c|}{COCO} & \multicolumn{3}{c|}{COCO} \\
    \rowcolor{mygray} dimension & (M)  & AP$^{\mathrm{box}}$$\uparrow$ &AP$^{\mathrm{box}}_{50}$$\uparrow$  & AP$^{\mathrm{box}}_{75}$$\uparrow$ & AP$^{\mathrm{box}}_s\uparrow$ & AP$^{\mathrm{box}}_m\uparrow$ & AP$^{\mathrm{box}}_l\uparrow$ & AP$^{\mathrm{mask}}$$\uparrow$ & AP$^{\mathrm{mask}}_{50}$$\uparrow$ & AP$^{\mathrm{mask}}_{75}$$\uparrow$ \\
    \hline\hline
    (48, 48, 96, 96) & 23.8M & 34.9 & 57.2  & 36.9 & 20.3 & 36.9 & 45.8 & 33.4  & 54.5 & 35.0\\
    (72, 72, 144, 144) & 24.0M & 35.2 & 57.2  & 37.7 & 20.1 & 37.4 & 46.7 & 33.5  & 54.5 & 35.6\\
    \textcolor{red}{(96, 96, 192, 192)} & 24.3M & 35.6 & 57.5  & 38.2 & 20.9 & 37.7 & 46.8 & 33.6  & 54.7 & 35.7\\
    (120, 120, 240, 240) & 24.5M & 35.5 & 57.5  & 37.8 & 20.7 & 37.6 & 47.4 & 33.8  & 54.7 & 35.9\\
    \hline
\end{tabular}}
\vspace{4pt}
\centering
\small
\caption{Number of channels in \texttt{encode} (Eq.~\ref{eq:center_init})}
\label{tab:ab_det}
\end{subtable}

\vspace{-10pt}
\end{table*}

\subsection{Experiments on Semantic Segmentation}
\label{sec:exp_sem_seg}

\noindent\textbf{Dataset.}$_{\!}$
ADE20K$_{\!}$~\citep{zhou2017scene} is a prominent semantic segmentation dataset$_{\!}$ renowned$_{\!}$ for$_{\!}$ its$_{\!}$ extensive$_{\!}$ collection$_{\!}$ of$_{\!}$ images. It$_{\!}$ encompasses$_{\!}$ 150$_{\!}$ object$_{\!}$ categories$_{\!}$ with$_{\!}$ a$_{\!}$ total$_{\!}$ of 25,000 images (20K/2K/3K for \texttt{train}/\texttt{val}/\texttt{test}).

\noindent\textbf{Training.$_{\!}$}
We$_{\!}$ use$_{\!}$ \textit{mmsegmentation}$_{\!}$~\citep{mmseg2020}$_{\!}$ as$_{\!}$ our$_{\!}$ codebase$_{\!}$ and follow the standard training protocols as detailed in~\citep{wang2021pyramid,ma2023image}. We evaluate FEC on a classic segmentation method, \ie, Semantic FPN~\citep{kirillov2019panoptic}, for its high efficiency. We utilize an AdamW~\citep{loshchilov2017decoupled} optimizer for 80K iterations using the polynomial decay learning rate scheduler with a power of 0.9. We set the batch size as 16 and the initial learning rate as 0.0001. More details are left in the appendix.

\noindent\textbf{Test.}
We use one input image scale with a shorter side of 512 during inference without applying data augmentation.

\noindent\textbf{Metric.} Mean intersection-over-union (mIoU) is reported.

\noindent\textbf{Performance$_{\!}$ Comparison.$_{\!}$}
From table~\ref{tab:ade20k} we can observe that our approach yields remarkable performance on dense prediction tasks. For example, our FEC-Base outperforms ResNet18, ResNet50, and PVT-Tiny by \textbf{4.8}\%, \textbf{1.0}\%, and \textbf{2.0}\% in terms of mIoU, respectively. In addition, FEC-Small can even outperform ResNet18, \ie, \textbf{35.3}\% \textit{vs} \textbf{32.9}\% in terms of mIoU. Such performance gains remain consistent as in the classification task and are significant considering the number of used parameters.

\subsection{Experiments on Object Detection and Instance Segmentation}
\label{sec:exp_det}
 
\noindent\textbf{Dataset.}$_{\!}$
The$_{\!}$ MS$_{\!}$ COCO$_{\!}$ 2017$_{\!}$ benchmark\citep{lin2014microsoft}$_{\!}$ features$_{\!}$ a di- verse$_{\!}$ collection$_{\!}$ of$_{\!}$ over$_{\!}$ 200K$_{\!}$ high-quality images,$_{\!}$ annotated across 80 common object categories in daily contexts.$_{\!}$ It$_{\!}$ is$_{\!}$ divided$_{\!}$ into$_{\!}$ 118K/5K/41K images$_{\!}$ for$_{\!}$ \texttt{train}/\texttt{val}/\texttt{test}.

\noindent\textbf{Training.}
We use \textit{mmdetection}~\citep{chen2019mmdetection} as our codebase and follow the training protocols as in~\citep{ma2023image}. We verify the effectiveness of FEC backbones on top of a milestone model, \ie, Mask R-CNN~\citep{he2017mask}. AdamW~\citep{loshchilov2017decoupled} optimizer is used for 12 epochs (1$\times$ scheduler) and initialize the backbone with ImageNet~\citep{ImageNet} pre-trained weights. A batch size of 16 and an initial learning rate of 0.0002 are used. More details are left in the appendix.

\noindent\textbf{Test.}
We use one input image scale with a shorter side of 800 during inference without applying data augmentation.

\noindent\textbf{Metric.} We report average precision (AP), AP$_{50}$, AP$_{75}$ for both object detection and instance segmentation. %, differentiated by superscripts (box or mask).

\noindent\textbf{Performance$_{\!}$ Comparison.$_{\!}$} 
Table~\ref{tab:det_seg}$_{\!}$ confirms$_{\!}$ again$_{\!}$ the$_{\!}$ tran-sferability and versatility of FEC for the common instance-centric recognition tasks. On top of a relatively conservative baseline, \ie, Mask-RCNN~\citep{he2017mask}, our algorithm outperforms both types of rivals. For instance, the performance of FEC-Tiny is clear ahead compared to ResNet-18~\citep{he2016deep} (\ie, \textbf{35.6}\% AP$^{\mathrm{box}}$ \textit{vs} 34.0\% AP$^{\mathrm{box}}$ and \textbf{33.6}\% AP$^{\mathrm{mask}}$ \textit{vs} 31.2\% AP$^{\mathrm{mask}}$), and FEC-Base achieves promising gains of \textbf{1.2}\% AP$^{\mathrm{box}}$ and \textbf{0.4}\% AP$^{\mathrm{mask}}$ against PVT-Tiny~\citep{wang2021pyramid}.

\subsection{Diagnostic Experiment}
\label{sec:diag_exp}

This$_{\!}$ section$_{\!}$ ablates$_{\!}$ FEC's$_{\!}$ key$_{\!}$ components$_{\!}$ on$_{\!}$ ImageNet$_{\!}$~\citep{ImageNet}$_{\!}$ \texttt{val}, ADE20K~\citep{zhou2017scene} \texttt{val}, and COCO~\citep{lin2014microsoft} \texttt{val2017}. All experiments use the FEC-Small model.

\noindent\textbf{Similarity$_{\!}$ Measurement.$_{\!}$}
We$_{\!}$ first$_{\!}$ examine$_{\!}$ the$_{\!}$ similarity measurement in the clustering process (Eq.~\ref{eq:compute_sim}) by contras- ting it with several standard similarity (distance) functions, \ie, Euclidean distance and dot product (unnormalized cosine similarity). As shown in Table~\ref{tab:ab_sim}, dot product and cosine similarity shows better performance than Euclidean distance, \eg, \textbf{72.7}\% \textit{vs} 72.2\% in terms of Top-1 accuracy. Cosine similarity is adopted by default.

\noindent\textbf{Feature$_{\!}$ Dispatching.$_{\!}$}
We$_{\!}$ then$_{\!}$ investigate$_{\!}$ the$_{\!}$ effects$_{\!}$ of$_{\!}$ the$_{\!}$ learnable parameters for similarity scaling and shifting (Eq.~\ref{eq:feat_dis}). In table~\ref{tab:ab_feat_dis}, the first three lines indicate the removal of the corresponding hyperparameters. We find that introducing these two factors can bring minor performance improvements, \eg, \textbf{72.7}\% \textit{vs} 72.1\% in terms of Top-1 accuracy and \textbf{91.2}\% \textit{vs} 90.8\% in terms of Top-5 accuracy.

\noindent\textbf{Feature Dimension.}
Last, we ablate the feature dimension $C'$ in the encoding layer (Eq.~\ref{eq:center_init}). In the pooling layer, $C'$ is used to increase the number of channels for the next stage. In the encoding layer, $C'$ is used to control the complexity of the projected \textit{key} and \textit{value} space (the output channels can be further adjusted by the \texttt{MLP} during feature dispatching). The results are summarized in Table~\ref{tab:ab_seg} and Table~\ref{tab:ab_det}. The four values of each row correspond to the four encoding phases. In a nutshell, the performance for downstream tasks improves as the feature dimension increases, \eg, \textbf{35.5}\% \textit{vs} 34.9\% in terms of mIoU, \textbf{35.6}\% \textit{vs} 34.9\% in terms of AP$^{\mathrm{box}}$, and \textbf{33.8}\% \textit{vs} 33.4\% in terms of AP$^{\mathrm{mask}}$, at the expense of parameter growth. For a fair comparison with previous work~\citep{ma2023image}, the settings in the 3$^{rd}$ row are adopted.

\section{Conclusion and Discussion}
\label{sec:conclusion}
In machine vision, extracting powerful distributed represen- tations for visual data while preserving the interpretability and explicit modeling of data distribution, presents a perennial challenge. While the community has witnessed great strides in visual backbones, top-leading solutions remain bound to the computational confines of processing rectangular image patches --- a stark contrast to the pixel organization observed in human perception. This study represents a significant leap forward by reformulating feature extraction as representative selection, resulting in a transparent and interpretable feature extractor. Our goal is to pave the way for vision systems that not only excel in performance but also possess an intrinsic understanding of the underlying data distribution of visual scenes, thereby enhancing both trust and clarity in their application.

{
    \small
    \bibliographystyle{unsrtnat}
    \bibliography{main}

\begin{thebibliography}{141}
\providecommand{\natexlab}[1]{#1}
\providecommand{\url}[1]{\texttt{#1}}
\expandafter\ifx\csname urlstyle\endcsname\relax
  \providecommand{\doi}[1]{doi: #1}\else
  \providecommand{\doi}{doi: \begingroup \urlstyle{rm}\Url}\fi

\bibitem[Snyder and Qi(2004)]{snyder2004machine}
Wesley~E Snyder and Hairong Qi.
\newblock \emph{Machine vision}, volume~1.
\newblock Cambridge University Press, 2004.

\bibitem[Bishop(2006)]{bishop2006pattern}
Christopher~M Bishop.
\newblock Pattern recognition.
\newblock \emph{Machine learning}, 128\penalty0 (9), 2006.

\bibitem[LeCun et~al.(2015)LeCun, Bengio, and Hinton]{lecun2015deep}
Yann LeCun, Yoshua Bengio, and Geoffrey Hinton.
\newblock Deep learning.
\newblock \emph{nature}, 521\penalty0 (7553):\penalty0 436--444, 2015.

\bibitem[Ng and Henikoff(2003)]{ng2003sift}
Pauline~C Ng and Steven Henikoff.
\newblock Sift: Predicting amino acid changes that affect protein function.
\newblock \emph{Nucleic acids research}, 31\penalty0 (13):\penalty0 3812--3814, 2003.

\bibitem[Dalal and Triggs(2005)]{dalal2005histograms}
Navneet Dalal and Bill Triggs.
\newblock Histograms of oriented gradients for human detection.
\newblock In \emph{CVPR}, pages 886--893, 2005.

\bibitem[Bay et~al.(2006)Bay, Tuytelaars, and Van~Gool]{bay2006surf}
Herbert Bay, Tinne Tuytelaars, and Luc Van~Gool.
\newblock Surf: Speeded up robust features.
\newblock In \emph{ECCV}, pages 404--417, 2006.

\bibitem[Calonder et~al.(2010)Calonder, Lepetit, Strecha, and Fua]{calonder2010brief}
Michael Calonder, Vincent Lepetit, Christoph Strecha, and Pascal Fua.
\newblock Brief: Binary robust independent elementary features.
\newblock In \emph{ECCV}, pages 778--792, 2010.

\bibitem[Rublee et~al.(2011)Rublee, Rabaud, Konolige, and Bradski]{rublee2011orb}
Ethan Rublee, Vincent Rabaud, Kurt Konolige, and Gary Bradski.
\newblock Orb: An efficient alternative to sift or surf.
\newblock In \emph{ICCV}, pages 2564--2571, 2011.

\bibitem[Leutenegger et~al.(2011)Leutenegger, Chli, and Siegwart]{leutenegger2011brisk}
Stefan Leutenegger, Margarita Chli, and Roland~Y Siegwart.
\newblock Brisk: Binary robust invariant scalable keypoints.
\newblock In \emph{ICCV}, pages 2548--2555, 2011.

\bibitem[Simonyan and Zisserman(2015)]{Simonyan15}
Karen Simonyan and Andrew Zisserman.
\newblock Very deep convolutional networks for large-scale image recognition.
\newblock In \emph{ICLR}, 2015.

\bibitem[He et~al.(2016)He, Zhang, Ren, and Sun]{he2016deep}
Kaiming He, Xiangyu Zhang, Shaoqing Ren, and Jian Sun.
\newblock Deep residual learning for image recognition.
\newblock In \emph{CVPR}, 2016.

\bibitem[Vaswani et~al.(2017)Vaswani, Shazeer, Parmar, Uszkoreit, Jones, Gomez, Kaiser, and Polosukhin]{vaswani2017attention}
Ashish Vaswani, Noam Shazeer, Niki Parmar, Jakob Uszkoreit, Llion Jones, Aidan~N Gomez, Lukasz Kaiser, and Illia Polosukhin.
\newblock Attention is all you need.
\newblock In \emph{NeurIPS}, 2017.

\bibitem[Dosovitskiy et~al.(2020)Dosovitskiy, Beyer, Kolesnikov, Weissenborn, Zhai, Unterthiner, Dehghani, Minderer, Heigold, Gelly, et~al.]{dosovitskiy2020image}
Alexey Dosovitskiy, Lucas Beyer, Alexander Kolesnikov, Dirk Weissenborn, Xiaohua Zhai, Thomas Unterthiner, Mostafa Dehghani, Matthias Minderer, Georg Heigold, Sylvain Gelly, et~al.
\newblock An image is worth 16x16 words: Transformers for image recognition at scale.
\newblock In \emph{ICLR}, 2020.

\bibitem[Tolstikhin et~al.(2021)Tolstikhin, Houlsby, Kolesnikov, Beyer, Zhai, Unterthiner, Yung, Steiner, Keysers, Uszkoreit, et~al.]{tolstikhin2021mlp}
Ilya~O Tolstikhin, Neil Houlsby, Alexander Kolesnikov, Lucas Beyer, Xiaohua Zhai, Thomas Unterthiner, Jessica Yung, Andreas Steiner, Daniel Keysers, Jakob Uszkoreit, et~al.
\newblock Mlp-mixer: An all-mlp architecture for vision.
\newblock In \emph{NeurIPS}, pages 24261--24272, 2021.

\bibitem[Touvron et~al.(2022)Touvron, Bojanowski, Caron, Cord, El-Nouby, Grave, Izacard, Joulin, Synnaeve, Verbeek, et~al.]{touvron2022resmlp}
Hugo Touvron, Piotr Bojanowski, Mathilde Caron, Matthieu Cord, Alaaeldin El-Nouby, Edouard Grave, Gautier Izacard, Armand Joulin, Gabriel Synnaeve, Jakob Verbeek, et~al.
\newblock Resmlp: Feedforward networks for image classification with data-efficient training.
\newblock \emph{IEEE TPAMI}, 45\penalty0 (4):\penalty0 5314--5321, 2022.

\bibitem[Gonzales and Wintz(1987)]{gonzales1987digital}
Rafael~C Gonzales and Paul Wintz.
\newblock \emph{Digital image processing}.
\newblock Addison-Wesley Longman Publishing Co., Inc., 1987.

\bibitem[Castleman(1996)]{castleman1996digital}
Kenneth~R Castleman.
\newblock \emph{Digital image processing}.
\newblock Prentice Hall Press, 1996.

\bibitem[Beymer and Poggio(1996)]{beymer1996image}
David Beymer and Tomaso Poggio.
\newblock Image representations for visual learning.
\newblock \emph{Science}, 272\penalty0 (5270):\penalty0 1905--1909, 1996.

\bibitem[Guidotti et~al.(2018)Guidotti, Monreale, Ruggieri, Turini, Giannotti, and Pedreschi]{guidotti2018survey}
Riccardo Guidotti, Anna Monreale, Salvatore Ruggieri, Franco Turini, Fosca Giannotti, and Dino Pedreschi.
\newblock A survey of methods for explaining black box models.
\newblock \emph{ACM computing surveys (CSUR)}, 51\penalty0 (5):\penalty0 1--42, 2018.

\bibitem[Weinlich et~al.(2016)Weinlich, Amon, Hutter, and Kaup]{weinlich2016probability}
Andreas Weinlich, Peter Amon, Andreas Hutter, and Andre Kaup.
\newblock Probability distribution estimation for autoregressive pixel-predictive image coding.
\newblock \emph{IEEE TIP}, 25\penalty0 (3):\penalty0 1382--1395, 2016.

\bibitem[Biederman(1987)]{biederman1987recognition}
Irving Biederman.
\newblock Recognition-by-components: a theory of human image understanding.
\newblock \emph{Psychological review}, 94\penalty0 (2):\penalty0 115, 1987.

\bibitem[Beutter and Stone(2000)]{beutter2000motion}
Brent~R Beutter and Leland~S Stone.
\newblock Motion coherence affects human perception and pursuit similarly.
\newblock \emph{Visual neuroscience}, 17\penalty0 (1):\penalty0 139--153, 2000.

\bibitem[Bill et~al.(2020)Bill, Pailian, Gershman, and Drugowitsch]{bill2020hierarchical}
Johannes Bill, Hrag Pailian, Samuel~J Gershman, and Jan Drugowitsch.
\newblock Hierarchical structure is employed by humans during visual motion perception.
\newblock \emph{Proceedings of the National Academy of Sciences}, 117\penalty0 (39):\penalty0 24581--24589, 2020.

\bibitem[Russakovsky et~al.(2015)Russakovsky, Deng, Su, Krause, Satheesh, Ma, Huang, Karpathy, Khosla, Bernstein, Berg, and Li]{ImageNet}
Olga Russakovsky, Jia Deng, Hao Su, Jonathan Krause, Sanjeev Satheesh, Sean Ma, Zhiheng Huang, Andrej Karpathy, Aditya Khosla, Michael~S. Bernstein, Alexander~C. Berg, and Fei{-}Fei Li.
\newblock Imagenet large scale visual recognition challenge.
\newblock \emph{IJCV}, 115\penalty0 (3):\penalty0 211--252, 2015.

\bibitem[Rommers et~al.(2013)Rommers, Dijkstra, and Bastiaansen]{rommers2013context}
Joost Rommers, Ton Dijkstra, and Marcel Bastiaansen.
\newblock Context-dependent semantic processing in the human brain: Evidence from idiom comprehension.
\newblock \emph{Journal of Cognitive Neuroscience}, 25\penalty0 (5):\penalty0 762--776, 2013.

\bibitem[Ma et~al.(2023)Ma, Zhou, Wang, Qin, Sun, Liu, and Fu]{ma2023image}
Xu~Ma, Yuqian Zhou, Huan Wang, Can Qin, Bin Sun, Chang Liu, and Yun Fu.
\newblock Image as set of points.
\newblock In \emph{ICLR}, 2023.

\bibitem[Liang et~al.(2023{\natexlab{a}})Liang, Cui, Wang, Geng, Wang, and Liu]{liang2023clusterfomer}
James Liang, Yiming Cui, Qifan Wang, Tong Geng, Wenguan Wang, and Dongfang Liu.
\newblock Clusterfomer: Clustering as a universal visual learner.
\newblock In \emph{NeurIPS}, 2023{\natexlab{a}}.

\bibitem[Selvaraju et~al.(2017)Selvaraju, Cogswell, Das, Vedantam, Parikh, and Batra]{selvaraju2017grad}
Ramprasaath~R Selvaraju, Michael Cogswell, Abhishek Das, Ramakrishna Vedantam, Devi Parikh, and Dhruv Batra.
\newblock Grad-cam: Visual explanations from deep networks via gradient-based localization.
\newblock In \emph{ICCV}, 2017.

\bibitem[Rudin et~al.(2022)Rudin, Chen, Chen, Huang, Semenova, and Zhong]{rudin2022interpretable}
Cynthia Rudin, Chaofan Chen, Zhi Chen, Haiyang Huang, Lesia Semenova, and Chudi Zhong.
\newblock Interpretable machine learning: Fundamental principles and 10 grand challenges.
\newblock \emph{Statistics Surveys}, 16:\penalty0 1--85, 2022.

\bibitem[Redmon et~al.(2016)Redmon, Divvala, Girshick, and Farhadi]{redmon2016you}
Joseph Redmon, Santosh Divvala, Ross Girshick, and Ali Farhadi.
\newblock You only look once: Unified, real-time object detection.
\newblock In \emph{CVPR}, pages 779--788, 2016.

\bibitem[Ren et~al.(2015)Ren, He, Girshick, and Sun]{ren2015faster}
Shaoqing Ren, Kaiming He, Ross Girshick, and Jian Sun.
\newblock Faster r-cnn: Towards real-time object detection with region proposal networks.
\newblock In \emph{NeurIPS}, 2015.

\bibitem[Carion et~al.(2020)Carion, Massa, Synnaeve, Usunier, Kirillov, and Zagoruyko]{carion2020end}
Nicolas Carion, Francisco Massa, Gabriel Synnaeve, Nicolas Usunier, Alexander Kirillov, and Sergey Zagoruyko.
\newblock End-to-end object detection with transformers.
\newblock In \emph{ECCV}, 2020.

\bibitem[Li et~al.(2022{\natexlab{a}})Li, Chen, Shi, Zhang, Yang, Liu, and Xiao]{li2022nicest}
Lin Li, Long Chen, Hanrong Shi, Hanwang Zhang, Yi~Yang, Wei Liu, and Jun Xiao.
\newblock Nicest: Noisy label correction and training for robust scene graph generation.
\newblock \emph{arXiv preprint arXiv:2207.13316}, 2022{\natexlab{a}}.

\bibitem[Li et~al.(2022{\natexlab{b}})Li, Chen, Huang, Zhang, Zhang, and Xiao]{li2022devil}
Lin Li, Long Chen, Yifeng Huang, Zhimeng Zhang, Songyang Zhang, and Jun Xiao.
\newblock The devil is in the labels: Noisy label correction for robust scene graph generation.
\newblock In \emph{CVPR}, pages 18869--18878, 2022{\natexlab{b}}.

\bibitem[Chen et~al.(2023)Chen, Li, Luo, and Xiao]{chen2023addressing}
Guikun Chen, Lin Li, Yawei Luo, and Jun Xiao.
\newblock Addressing predicate overlap in scene graph generation with semantic granularity controller.
\newblock In \emph{ICME}, pages 78--83, 2023.

\bibitem[Li et~al.(2023{\natexlab{a}})Li, Chen, Xiao, Yang, Wang, and Chen]{li2023compositional}
Lin Li, Guikun Chen, Jun Xiao, Yi~Yang, Chunping Wang, and Long Chen.
\newblock Compositional feature augmentation for unbiased scene graph generation.
\newblock In \emph{ICCV}, pages 21685--21695, 2023{\natexlab{a}}.

\bibitem[Yin et~al.(2022)Yin, Zhou, Zhang, Fang, Xu, Shen, and Wang]{yin2022proposalcontrast}
Junbo Yin, Dingfu Zhou, Liangjun Zhang, Jin Fang, Cheng-Zhong Xu, Jianbing Shen, and Wenguan Wang.
\newblock Proposalcontrast: Unsupervised pre-training for lidar-based 3d object detection.
\newblock In \emph{ECCV}, pages 17--33, 2022.

\bibitem[Feng et~al.(2023)Feng, Wang, Wang, Yang, and Zheng]{feng2023clustering}
Tuo Feng, Wenguan Wang, Xiaohan Wang, Yi~Yang, and Qinghua Zheng.
\newblock Clustering based point cloud representation learning for 3d analysis.
\newblock In \emph{ICCV}, pages 8283--8294, 2023.

\bibitem[Feng et~al.(2024)Feng, Quan, Wang, Wang, and Yang]{feng2024interpretable3D}
Tuo Feng, Ruijie Quan, Xiaohan Wang, Wenguan Wang, and Yi~Yang.
\newblock Interpretable3d: An ad-hoc interpretable classifier for 3d point clouds.
\newblock In \emph{AAAI}, 2024.

\bibitem[Wang et~al.(2018)Wang, Shen, Porikli, and Yang]{wang2018semi}
Wenguan Wang, Jianbing Shen, Fatih Porikli, and Ruigang Yang.
\newblock Semi-supervised video object segmentation with super-trajectories.
\newblock \emph{IEEE TPAMI}, 41\penalty0 (4):\penalty0 985--998, 2018.

\bibitem[Liang et~al.(2022)Liang, Wang, Miao, and Yang]{liang2022gmmseg}
Chen Liang, Wenguan Wang, Jiaxu Miao, and Yi~Yang.
\newblock Gmmseg: Gaussian mixture based generative semantic segmentation models.
\newblock In \emph{NeurIPS}, pages 31360--31375, 2022.

\bibitem[Zhou et~al.(2022)Zhou, Wang, Konukoglu, and Van~Gool]{zhou2022rethinking}
Tianfei Zhou, Wenguan Wang, Ender Konukoglu, and Luc Van~Gool.
\newblock Rethinking semantic segmentation: A prototype view.
\newblock In \emph{CVPR}, 2022.

\bibitem[Liang et~al.(2023{\natexlab{b}})Liang, Zhou, Liu, and Wang]{liang2023clustseg}
James~Chenhao Liang, Tianfei Zhou, Dongfang Liu, and Wenguan Wang.
\newblock Clustseg: Clustering for universal segmentation.
\newblock In \emph{ICML}, pages 20787--20809, 2023{\natexlab{b}}.

\bibitem[Li et~al.(2023{\natexlab{b}})Li, Wang, Zhou, Li, and Yang]{li2023unified}
Liulei Li, Wenguan Wang, Tianfei Zhou, Jianwu Li, and Yi~Yang.
\newblock Unified mask embedding and correspondence learning for self-supervised video segmentation.
\newblock In \emph{CVPR}, pages 18706--18716, 2023{\natexlab{b}}.

\bibitem[Grabski et~al.(2023)Grabski, Street, and Irizarry]{grabski2023significance}
Isabella~N Grabski, Kelly Street, and Rafael~A Irizarry.
\newblock Significance analysis for clustering with single-cell rna-sequencing data.
\newblock \emph{Nature Methods}, 20\penalty0 (8):\penalty0 1196--1202, 2023.

\bibitem[Barrio-Hernandez et~al.(2023)Barrio-Hernandez, Yeo, J{\"a}nes, Mirdita, Gilchrist, Wein, Varadi, Velankar, Beltrao, and Steinegger]{barrio2023clustering}
Inigo Barrio-Hernandez, Jingi Yeo, J{\"u}rgen J{\"a}nes, Milot Mirdita, Cameron~LM Gilchrist, Tanita Wein, Mihaly Varadi, Sameer Velankar, Pedro Beltrao, and Martin Steinegger.
\newblock Clustering predicted structures at the scale of the known protein universe.
\newblock \emph{Nature}, 622\penalty0 (7983):\penalty0 637--645, 2023.

\bibitem[Wayment-Steele et~al.(2024)Wayment-Steele, Ojoawo, Otten, Apitz, Pitsawong, H{\"o}mberger, Ovchinnikov, Colwell, and Kern]{wayment2024predicting}
Hannah~K Wayment-Steele, Adedolapo Ojoawo, Renee Otten, Julia~M Apitz, Warintra Pitsawong, Marc H{\"o}mberger, Sergey Ovchinnikov, Lucy Colwell, and Dorothee Kern.
\newblock Predicting multiple conformations via sequence clustering and alphafold2.
\newblock \emph{Nature}, 625\penalty0 (7996):\penalty0 832--839, 2024.

\bibitem[Quan et~al.(2024)Quan, Wang, Ma, Fan, and Yang]{quan2024clustering}
Ruijie Quan, Wenguan Wang, Fan Ma, Hehe Fan, and Yi~Yang.
\newblock Clustering for protein representation learning.
\newblock In \emph{CVPR}, 2024.

\bibitem[Ren and Malik(2003)]{ren2003learning}
Ren and Malik.
\newblock Learning a classification model for segmentation.
\newblock In \emph{ICCV}, pages 10--17, 2003.

\bibitem[Jaegle et~al.(2021)Jaegle, Gimeno, Brock, Vinyals, Zisserman, and Carreira]{jaegle2021perceiver}
Andrew Jaegle, Felix Gimeno, Andy Brock, Oriol Vinyals, Andrew Zisserman, and Joao Carreira.
\newblock Perceiver: General perception with iterative attention.
\newblock In \emph{ICML}, pages 4651--4664, 2021.

\bibitem[Carreira et~al.(2022)Carreira, Koppula, Zoran, Recasens, Ionescu, Henaff, Shelhamer, Arandjelovic, Botvinick, Vinyals, et~al.]{carreira2022hip}
Joao Carreira, Skanda Koppula, Daniel Zoran, Adria Recasens, Catalin Ionescu, Olivier Henaff, Evan Shelhamer, Relja Arandjelovic, Matt Botvinick, Oriol Vinyals, et~al.
\newblock Hip: Hierarchical perceiver.
\newblock \emph{arXiv preprint arXiv:2202.10890}, 2022.

\bibitem[Gu et~al.(2009)Gu, Lim, Arbel{\'a}ez, and Malik]{gu2009recognition}
Chunhui Gu, Joseph~J Lim, Pablo Arbel{\'a}ez, and Jitendra Malik.
\newblock Recognition using regions.
\newblock In \emph{CVPR}, pages 1030--1037, 2009.

\bibitem[Carreira et~al.(2012)Carreira, Caseiro, Batista, and Sminchisescu]{carreira2012semantic}
Joao Carreira, Rui Caseiro, Jorge Batista, and Cristian Sminchisescu.
\newblock Semantic segmentation with second-order pooling.
\newblock In \emph{ECCV}, pages 430--443, 2012.

\bibitem[Krizhevsky et~al.(2012)Krizhevsky, Sutskever, and Hinton]{krizhevsky2012imagenet}
Alex Krizhevsky, Ilya Sutskever, and Geoffrey~E Hinton.
\newblock Imagenet classification with deep convolutional neural networks.
\newblock In \emph{NeurIPS}, 2012.

\bibitem[Liu et~al.(2022)Liu, Mao, Wu, Feichtenhofer, Darrell, and Xie]{liu2022convnet}
Zhuang Liu, Hanzi Mao, Chao-Yuan Wu, Christoph Feichtenhofer, Trevor Darrell, and Saining Xie.
\newblock A convnet for the 2020s.
\newblock In \emph{CVPR}, pages 11976--11986, 2022.

\bibitem[Szegedy et~al.(2015)Szegedy, Liu, Jia, Sermanet, Reed, Anguelov, Erhan, Vanhoucke, and Rabinovich]{szegedy2015going}
Christian Szegedy, Wei Liu, Yangqing Jia, Pierre Sermanet, Scott Reed, Dragomir Anguelov, Dumitru Erhan, Vincent Vanhoucke, and Andrew Rabinovich.
\newblock Going deeper with convolutions.
\newblock In \emph{CVPR}, pages 1--9, 2015.

\bibitem[Huang et~al.(2017)Huang, Liu, Van Der~Maaten, and Weinberger]{huang2017densely}
Gao Huang, Zhuang Liu, Laurens Van Der~Maaten, and Kilian~Q Weinberger.
\newblock Densely connected convolutional networks.
\newblock In \emph{CVPR}, pages 4700--4708, 2017.

\bibitem[Tan and Le(2019)]{tan2019efficientnet}
Mingxing Tan and Quoc Le.
\newblock Efficientnet: Rethinking model scaling for convolutional neural networks.
\newblock In \emph{ICML}, pages 6105--6114, 2019.

\bibitem[Howard et~al.(2017)Howard, Zhu, Chen, Kalenichenko, Wang, Weyand, Andreetto, and Adam]{howard2017mobilenets}
Andrew~G Howard, Menglong Zhu, Bo~Chen, Dmitry Kalenichenko, Weijun Wang, Tobias Weyand, Marco Andreetto, and Hartwig Adam.
\newblock Mobilenets: Efficient convolutional neural networks for mobile vision applications.
\newblock In \emph{CVPR}, 2017.

\bibitem[Dai et~al.(2017)Dai, Qi, Xiong, Li, Zhang, Hu, and Wei]{dai2017deformable}
Jifeng Dai, Haozhi Qi, Yuwen Xiong, Yi~Li, Guodong Zhang, Han Hu, and Yichen Wei.
\newblock Deformable convolutional networks.
\newblock In \emph{CVPR}, 2017.

\bibitem[Khan et~al.(2022)Khan, Naseer, Hayat, Zamir, Khan, and Shah]{khan2022transformers}
Salman Khan, Muzammal Naseer, Munawar Hayat, Syed~Waqas Zamir, Fahad~Shahbaz Khan, and Mubarak Shah.
\newblock Transformers in vision: A survey.
\newblock \emph{ACM computing surveys (CSUR)}, 54\penalty0 (10s):\penalty0 1--41, 2022.

\bibitem[Gani et~al.(2022)Gani, Naseer, and Yaqub]{gani2022train}
Hanan Gani, Muzammal Naseer, and Mohammad Yaqub.
\newblock How to train vision transformer on small-scale datasets?
\newblock In \emph{BMVC}, 2022.

\bibitem[Dai et~al.(2021)Dai, Liu, Le, and Tan]{dai2021coatnet}
Zihang Dai, Hanxiao Liu, Quoc~V Le, and Mingxing Tan.
\newblock Coatnet: Marrying convolution and attention for all data sizes.
\newblock In \emph{NeurIPS}, pages 3965--3977, 2021.

\bibitem[Chen et~al.(2022{\natexlab{a}})Chen, Dai, Chen, Liu, Dong, Yuan, and Liu]{chen2022mobile}
Yinpeng Chen, Xiyang Dai, Dongdong Chen, Mengchen Liu, Xiaoyi Dong, Lu~Yuan, and Zicheng Liu.
\newblock Mobile-former: Bridging mobilenet and transformer.
\newblock In \emph{CVPR}, pages 5270--5279, 2022{\natexlab{a}}.

\bibitem[Guo et~al.(2022)Guo, Tang, Han, Chen, Wu, Xu, Xu, and Wang]{guo2022hire}
Jianyuan Guo, Yehui Tang, Kai Han, Xinghao Chen, Han Wu, Chao Xu, Chang Xu, and Yunhe Wang.
\newblock Hire-mlp: Vision mlp via hierarchical rearrangement.
\newblock In \emph{CVPR}, pages 826--836, 2022.

\bibitem[Hou et~al.(2022)Hou, Jiang, Yuan, Cheng, Yan, and Feng]{hou2022vision}
Qibin Hou, Zihang Jiang, Li~Yuan, Ming-Ming Cheng, Shuicheng Yan, and Jiashi Feng.
\newblock Vision permutator: A permutable mlp-like architecture for visual recognition.
\newblock \emph{IEEE TPAMI}, 45\penalty0 (1):\penalty0 1328--1334, 2022.

\bibitem[Tang et~al.(2022)Tang, Han, Guo, Xu, Li, Xu, and Wang]{tang2022image}
Yehui Tang, Kai Han, Jianyuan Guo, Chang Xu, Yanxi Li, Chao Xu, and Yunhe Wang.
\newblock An image patch is a wave: Phase-aware vision mlp.
\newblock In \emph{CVPR}, pages 10935--10944, 2022.

\bibitem[Chen et~al.(2022{\natexlab{b}})Chen, Xie, Ge, Chen, Liang, and Luo]{chen2021cyclemlp}
Shoufa Chen, Enze Xie, Chongjian Ge, Runjian Chen, Ding Liang, and Ping Luo.
\newblock Cyclemlp: A mlp-like architecture for dense prediction.
\newblock In \emph{ICLR}, 2022{\natexlab{b}}.

\bibitem[Huang et~al.(2021)Huang, Ben, Luo, Cheng, Yu, and Fu]{huang2021shuffle}
Zilong Huang, Youcheng Ben, Guozhong Luo, Pei Cheng, Gang Yu, and Bin Fu.
\newblock Shuffle transformer: Rethinking spatial shuffle for vision transformer.
\newblock \emph{arXiv preprint arXiv:2106.03650}, 2021.

\bibitem[Lian et~al.(2021)Lian, Yu, Sun, and Gao]{lian2021mlp}
Dongze Lian, Zehao Yu, Xing Sun, and Shenghua Gao.
\newblock As-mlp: An axial shifted mlp architecture for vision.
\newblock In \emph{ICLR}, 2021.

\bibitem[Yu et~al.(2022)Yu, Luo, Zhou, Si, Zhou, Wang, Feng, and Yan]{yu2022metaformer}
Weihao Yu, Mi~Luo, Pan Zhou, Chenyang Si, Yichen Zhou, Xinchao Wang, Jiashi Feng, and Shuicheng Yan.
\newblock Metaformer is actually what you need for vision.
\newblock In \emph{CVPR}, pages 10819--10829, 2022.

\bibitem[Laugel et~al.(2019)Laugel, Lesot, Marsala, Renard, and Detyniecki]{laugel2019dangers}
Thibault Laugel, Marie-Jeanne Lesot, Christophe Marsala, Xavier Renard, and Marcin Detyniecki.
\newblock The dangers of post-hoc interpretability: Unjustified counterfactual explanations.
\newblock In \emph{IJCAI}, 2019.

\bibitem[Rudin(2019)]{rudin2019stop}
Cynthia Rudin.
\newblock Stop explaining black box machine learning models for high stakes decisions and use interpretable models instead.
\newblock \emph{Nature Machine Intelligence}, 1\penalty0 (5):\penalty0 206--215, 2019.

\bibitem[Arrieta et~al.(2020)Arrieta, D{\'\i}az-Rodr{\'\i}guez, Del~Ser, Bennetot, Tabik, Barbado, Garc{\'\i}a, Gil-L{\'o}pez, Molina, Benjamins, et~al.]{arrieta2020explainable}
Alejandro~Barredo Arrieta, Natalia D{\'\i}az-Rodr{\'\i}guez, Javier Del~Ser, Adrien Bennetot, Siham Tabik, Alberto Barbado, Salvador Garc{\'\i}a, Sergio Gil-L{\'o}pez, Daniel Molina, Richard Benjamins, et~al.
\newblock Explainable artificial intelligence (xai): Concepts, taxonomies, opportunities and challenges toward responsible ai.
\newblock \emph{Information Fusion}, 58:\penalty0 82--115, 2020.

\bibitem[Chen et~al.(2019{\natexlab{a}})Chen, Li, Tao, Barnett, Rudin, and Su]{chen2019looks}
Chaofan Chen, Oscar Li, Daniel Tao, Alina Barnett, Cynthia Rudin, and Jonathan~K Su.
\newblock This looks like that: deep learning for interpretable image recognition.
\newblock In \emph{NeurIPS}, 2019{\natexlab{a}}.

\bibitem[Erhan et~al.(2009)Erhan, Bengio, Courville, and Vincent]{erhan2009visualizing}
Dumitru Erhan, Yoshua Bengio, Aaron Courville, and Pascal Vincent.
\newblock Visualizing higher-layer features of a deep network.
\newblock \emph{University of Montreal}, 1341\penalty0 (3):\penalty0 1, 2009.

\bibitem[Simonyan et~al.(2013)Simonyan, Vedaldi, and Zisserman]{simonyan2013deep}
Karen Simonyan, Andrea Vedaldi, and Andrew Zisserman.
\newblock Deep inside convolutional networks: Visualising image classification models and saliency maps.
\newblock \emph{arXiv preprint arXiv:1312.6034}, 2013.

\bibitem[Zeiler and Fergus(2014)]{zeiler2014visualizing}
Matthew~D Zeiler and Rob Fergus.
\newblock Visualizing and understanding convolutional networks.
\newblock In \emph{ECCV}, 2014.

\bibitem[Mahendran and Vedaldi(2015)]{mahendran2015understanding}
Aravindh Mahendran and Andrea Vedaldi.
\newblock Understanding deep image representations by inverting them.
\newblock In \emph{CVPR}, 2015.

\bibitem[Yosinski et~al.(2015)Yosinski, Clune, Nguyen, Fuchs, and Lipson]{yosinski2015understanding}
Jason Yosinski, Jeff Clune, Anh Nguyen, Thomas Fuchs, and Hod Lipson.
\newblock Understanding neural networks through deep visualization.
\newblock \emph{arXiv preprint arXiv:1506.06579}, 2015.

\bibitem[Bach et~al.(2015)Bach, Binder, Montavon, Klauschen, M{\"u}ller, and Samek]{bach2015pixel}
Sebastian Bach, Alexander Binder, Gr{\'e}goire Montavon, Frederick Klauschen, Klaus-Robert M{\"u}ller, and Wojciech Samek.
\newblock On pixel-wise explanations for non-linear classifier decisions by layer-wise relevance propagation.
\newblock \emph{PloS one}, 10\penalty0 (7):\penalty0 e0130140, 2015.

\bibitem[Zhou et~al.(2016)Zhou, Khosla, Lapedriza, Oliva, and Torralba]{cam2016learning}
Bolei Zhou, Aditya Khosla, Agata Lapedriza, Aude Oliva, and Antonio Torralba.
\newblock Learning deep features for discriminative localization.
\newblock In \emph{CVPR}, 2016.

\bibitem[Shrikumar et~al.(2017)Shrikumar, Greenside, and Kundaje]{shrikumar2017learning}
Avanti Shrikumar, Peyton Greenside, and Anshul Kundaje.
\newblock Learning important features through propagating activation differences.
\newblock In \emph{ICML}, 2017.

\bibitem[Ribeiro et~al.(2016)Ribeiro, Singh, and Guestrin]{ribeiro2016should}
Marco~Tulio Ribeiro, Sameer Singh, and Carlos Guestrin.
\newblock "why should i trust you?" explaining the predictions of any classifier.
\newblock In \emph{KDD}, 2016.

\bibitem[Zintgraf et~al.(2017)Zintgraf, Cohen, Adel, and Welling]{zintgraf2017visualizing}
Luisa~M Zintgraf, Taco~S Cohen, Tameem Adel, and Max Welling.
\newblock Visualizing deep neural network decisions: Prediction difference analysis.
\newblock In \emph{ICLR}, 2017.

\bibitem[Koh and Liang(2017)]{koh2017understanding}
Pang~Wei Koh and Percy Liang.
\newblock Understanding black-box predictions via influence functions.
\newblock In \emph{ICML}, 2017.

\bibitem[Lundberg and Lee(2017)]{lundberg2017unified}
Scott~M Lundberg and Su-In Lee.
\newblock A unified approach to interpreting model predictions.
\newblock In \emph{NeurIPS}, 2017.

\bibitem[Caron et~al.(2021)Caron, Touvron, Misra, J{\'e}gou, Mairal, Bojanowski, and Joulin]{caron2021emerging}
Mathilde Caron, Hugo Touvron, Ishan Misra, Herv{\'e} J{\'e}gou, Julien Mairal, Piotr Bojanowski, and Armand Joulin.
\newblock Emerging properties in self-supervised vision transformers.
\newblock In \emph{ICCV}, pages 9650--9660, 2021.

\bibitem[Oquab et~al.(2023)Oquab, Darcet, Moutakanni, Vo, Szafraniec, Khalidov, Fernandez, Haziza, Massa, El-Nouby, et~al.]{oquab2023dinov2}
Maxime Oquab, Timoth{\'e}e Darcet, Th{\'e}o Moutakanni, Huy Vo, Marc Szafraniec, Vasil Khalidov, Pierre Fernandez, Daniel Haziza, Francisco Massa, Alaaeldin El-Nouby, et~al.
\newblock Dinov2: Learning robust visual features without supervision.
\newblock \emph{arXiv preprint arXiv:2304.07193}, 2023.

\bibitem[Alvarez~Melis and Jaakkola(2018)]{alvarez2018towards}
David Alvarez~Melis and Tommi Jaakkola.
\newblock Towards robust interpretability with self-explaining neural networks.
\newblock In \emph{NeurIPS}, 2018.

\bibitem[Kim et~al.(2018)Kim, Wattenberg, Gilmer, Cai, Wexler, Viegas, et~al.]{kim2018interpretability}
Been Kim, Martin Wattenberg, Justin Gilmer, Carrie Cai, James Wexler, Fernanda Viegas, et~al.
\newblock Interpretability beyond feature attribution: Quantitative testing with concept activation vectors (tcav).
\newblock In \emph{ICML}, 2018.

\bibitem[Wang(2019)]{wang2019gaining}
Tong Wang.
\newblock Gaining free or low-cost interpretability with interpretable partial substitute.
\newblock In \emph{ICML}, 2019.

\bibitem[Subramanian et~al.(2018)Subramanian, Pruthi, Jhamtani, Berg-Kirkpatrick, and Hovy]{subramanian2018spine}
Anant Subramanian, Danish Pruthi, Harsh Jhamtani, Taylor Berg-Kirkpatrick, and Eduard Hovy.
\newblock Spine: Sparse interpretable neural embeddings.
\newblock In \emph{AAAI}, 2018.

\bibitem[Chen et~al.(2016)Chen, Duan, Houthooft, Schulman, Sutskever, and Abbeel]{chen2016infogan}
Xi~Chen, Yan Duan, Rein Houthooft, John Schulman, Ilya Sutskever, and Pieter Abbeel.
\newblock Infogan: Interpretable representation learning by information maximizing generative adversarial nets.
\newblock In \emph{NeurIPS}, 2016.

\bibitem[You et~al.(2017)You, Ding, Canini, Pfeifer, and Gupta]{you2017deep}
Seungil You, David Ding, Kevin Canini, Jan Pfeifer, and Maya Gupta.
\newblock Deep lattice networks and partial monotonic functions.
\newblock In \emph{NeurIPS}, 2017.

\bibitem[Wang et~al.(2023)Wang, Han, Zhou, and Liu]{wang2023visual}
Wenguan Wang, Cheng Han, Tianfei Zhou, and Dongfang Liu.
\newblock Visual recognition with deep nearest centroids.
\newblock In \emph{ICLR}, 2023.

\bibitem[Yu et~al.(2023{\natexlab{a}})Yu, Chu, Tong, Wu, Pai, Buchanan, and Ma]{yu2023emergence}
Yaodong Yu, Tianzhe Chu, Shengbang Tong, Ziyang Wu, Druv Pai, Sam Buchanan, and Yi~Ma.
\newblock Emergence of segmentation with minimalistic white-box transformers.
\newblock \emph{arXiv preprint arXiv:2308.16271}, 2023{\natexlab{a}}.

\bibitem[Yu et~al.(2023{\natexlab{b}})Yu, Buchanan, Pai, Chu, Wu, Tong, Haeffele, and Ma]{yu2023white}
Yaodong Yu, Sam Buchanan, Druv Pai, Tianzhe Chu, Ziyang Wu, Shengbang Tong, Benjamin~D Haeffele, and Yi~Ma.
\newblock White-box transformers via sparse rate reduction.
\newblock \emph{arXiv preprint arXiv:2306.01129}, 2023{\natexlab{b}}.

\bibitem[Lin et~al.(2014)Lin, Maire, Belongie, Hays, Perona, Ramanan, Doll{\'a}r, and Zitnick]{lin2014microsoft}
Tsung-Yi Lin, Michael Maire, Serge Belongie, James Hays, Pietro Perona, Deva Ramanan, Piotr Doll{\'a}r, and C~Lawrence Zitnick.
\newblock Microsoft coco: Common objects in context.
\newblock In \emph{ECCV}, 2014.

\bibitem[Zhou et~al.(2017)Zhou, Zhao, Puig, Fidler, Barriuso, and Torralba]{zhou2017scene}
Bolei Zhou, Hang Zhao, Xavier Puig, Sanja Fidler, Adela Barriuso, and Antonio Torralba.
\newblock Scene parsing through ade20k dataset.
\newblock In \emph{CVPR}, 2017.

\bibitem[Wightman(2019)]{rw2019timm}
Ross Wightman.
\newblock Pytorch image models.
\newblock \url{https://github.com/rwightman/pytorch-image-models}, 2019.

\bibitem[Loshchilov and Hutter(2017)]{loshchilov2017decoupled}
Ilya Loshchilov and Frank Hutter.
\newblock Decoupled weight decay regularization.
\newblock \emph{arXiv preprint arXiv:1711.05101}, 2017.

\bibitem[Liu et~al.(2021{\natexlab{a}})Liu, Lin, Cao, Hu, Wei, Zhang, Lin, and Guo]{liu2021swin}
Ze~Liu, Yutong Lin, Yue Cao, Han Hu, Yixuan Wei, Zheng Zhang, Stephen Lin, and Baining Guo.
\newblock Swin transformer: Hierarchical vision transformer using shifted windows.
\newblock In \emph{ICCV}, 2021{\natexlab{a}}.

\bibitem[Touvron et~al.(2021)Touvron, Cord, Douze, Massa, Sablayrolles, and J{\'e}gou]{touvron2021training}
Hugo Touvron, Matthieu Cord, Matthijs Douze, Francisco Massa, Alexandre Sablayrolles, and Herv{\'e} J{\'e}gou.
\newblock Training data-efficient image transformers \& distillation through attention.
\newblock In \emph{ICML}, pages 10347--10357, 2021.

\bibitem[Liu et~al.(2021{\natexlab{b}})Liu, Dai, So, and Le]{liu2021pay}
Hanxiao Liu, Zihang Dai, David So, and Quoc~V Le.
\newblock Pay attention to mlps.
\newblock In \emph{NeurIPS}, pages 9204--9215, 2021{\natexlab{b}}.

\bibitem[Wang et~al.(2021{\natexlab{a}})Wang, Xie, Li, Fan, Song, Liang, Lu, Luo, and Shao]{wang2021pyramid}
Wenhai Wang, Enze Xie, Xiang Li, Deng-Ping Fan, Kaitao Song, Ding Liang, Tong Lu, Ping Luo, and Ling Shao.
\newblock Pyramid vision transformer: A versatile backbone for dense prediction without convolutions.
\newblock In \emph{ICCV}, pages 568--578, 2021{\natexlab{a}}.

\bibitem[Trockman and Kolter(2022)]{trockman2022patches}
Asher Trockman and J~Zico Kolter.
\newblock Patches are all you need?
\newblock \emph{arXiv preprint arXiv:2201.09792}, 2022.

\bibitem[Kirillov et~al.(2019)Kirillov, Girshick, He, and Doll{\'a}r]{kirillov2019panoptic}
Alexander Kirillov, Ross Girshick, Kaiming He, and Piotr Doll{\'a}r.
\newblock Panoptic feature pyramid networks.
\newblock In \emph{CVPR}, pages 6399--6408, 2019.

\bibitem[He et~al.(2017)He, Gkioxari, Doll{\'a}r, and Girshick]{he2017mask}
Kaiming He, Georgia Gkioxari, Piotr Doll{\'a}r, and Ross Girshick.
\newblock Mask r-cnn.
\newblock In \emph{ICCV}, pages 2961--2969, 2017.

\bibitem[Pedregosa et~al.(2011)Pedregosa, Varoquaux, Gramfort, Michel, Thirion, Grisel, Blondel, Prettenhofer, Weiss, Dubourg, Vanderplas, Passos, Cournapeau, Brucher, Perrot, and Duchesnay]{scikit-learn}
F.~Pedregosa, G.~Varoquaux, A.~Gramfort, V.~Michel, B.~Thirion, O.~Grisel, M.~Blondel, P.~Prettenhofer, R.~Weiss, V.~Dubourg, J.~Vanderplas, A.~Passos, D.~Cournapeau, M.~Brucher, M.~Perrot, and E.~Duchesnay.
\newblock Scikit-learn: Machine learning in {P}ython.
\newblock \emph{JMLR}, 12:\penalty0 2825--2830, 2011.

\bibitem[Contributors(2020)]{mmseg2020}
MMSegmentation Contributors.
\newblock {MMSegmentation}: Openmmlab semantic segmentation toolbox and benchmark.
\newblock \url{https://github.com/open-mmlab/mmsegmentation}, 2020.

\bibitem[Chen et~al.(2019{\natexlab{b}})Chen, Wang, Pang, Cao, Xiong, Li, Sun, Feng, Liu, Xu, et~al.]{chen2019mmdetection}
Kai Chen, Jiaqi Wang, Jiangmiao Pang, Yuhang Cao, Yu~Xiong, Xiaoxiao Li, Shuyang Sun, Wansen Feng, Ziwei Liu, Jiarui Xu, et~al.
\newblock Mmdetection: Open mmlab detection toolbox and benchmark.
\newblock \emph{arXiv preprint arXiv:1906.07155}, 2019{\natexlab{b}}.

\bibitem[Zhong et~al.(2020)Zhong, Zheng, Kang, Li, and Yang]{zhong2020random}
Zhun Zhong, Liang Zheng, Guoliang Kang, Shaozi Li, and Yi~Yang.
\newblock Random erasing data augmentation.
\newblock In \emph{AAAI}, pages 13001--13008, 2020.

\bibitem[Zhang et~al.(2017)Zhang, Cisse, Dauphin, and Lopez-Paz]{zhang2017mixup}
Hongyi Zhang, Moustapha Cisse, Yann~N Dauphin, and David Lopez-Paz.
\newblock mixup: Beyond empirical risk minimization.
\newblock \emph{arXiv preprint arXiv:1710.09412}, 2017.

\bibitem[Yun et~al.(2019)Yun, Han, Oh, Chun, Choe, and Yoo]{yun2019cutmix}
Sangdoo Yun, Dongyoon Han, Seong~Joon Oh, Sanghyuk Chun, Junsuk Choe, and Youngjoon Yoo.
\newblock Cutmix: Regularization strategy to train strong classifiers with localizable features.
\newblock In \emph{ICCV}, pages 6023--6032, 2019.

\bibitem[Szegedy et~al.(2016)Szegedy, Vanhoucke, Ioffe, Shlens, and Wojna]{szegedy2016rethinking}
Christian Szegedy, Vincent Vanhoucke, Sergey Ioffe, Jon Shlens, and Zbigniew Wojna.
\newblock Rethinking the inception architecture for computer vision.
\newblock In \emph{CVPR}, pages 2818--2826, 2016.

\bibitem[Polyak and Juditsky(1992)]{polyak1992acceleration}
Boris~T Polyak and Anatoli~B Juditsky.
\newblock Acceleration of stochastic approximation by averaging.
\newblock \emph{SIAM journal on control and optimization}, 30\penalty0 (4):\penalty0 838--855, 1992.

\bibitem[Mitchell(1999)]{mitchell1999machine}
Tom~M Mitchell.
\newblock Machine learning and data mining.
\newblock \emph{Communications of the ACM}, 42\penalty0 (11):\penalty0 30--36, 1999.

\bibitem[Xu and Wunsch(2005)]{xu2005survey}
Rui Xu and Donald Wunsch.
\newblock Survey of clustering algorithms ieee transactions on neural networks, vol. 16 (3), 2005.

\bibitem[Van~Gansbeke et~al.(2020)Van~Gansbeke, Vandenhende, Georgoulis, Proesmans, and Van~Gool]{van2020scan}
Wouter Van~Gansbeke, Simon Vandenhende, Stamatios Georgoulis, Marc Proesmans, and Luc Van~Gool.
\newblock Scan: Learning to classify images without labels.
\newblock In \emph{ECCV}, 2020.

\bibitem[Duncan and Ayache(2000)]{duncan2000medical}
James~S Duncan and Nicholas Ayache.
\newblock Medical image analysis: Progress over two decades and the challenges ahead.
\newblock \emph{TPAMI}, 22\penalty0 (1):\penalty0 85--106, 2000.

\bibitem[Yang et~al.(2020)Yang, Wei, Zhang, and Zhu]{yang2020design}
Xiao Yang, Fangyun Wei, Hongyang Zhang, and Jun Zhu.
\newblock Design and interpretation of universal adversarial patches in face detection.
\newblock In \emph{ECCV}, pages 174--191, 2020.

\bibitem[Dong et~al.(2019)Dong, Su, Wu, Li, Liu, Zhang, and Zhu]{dong2019efficient}
Yinpeng Dong, Hang Su, Baoyuan Wu, Zhifeng Li, Wei Liu, Tong Zhang, and Jun Zhu.
\newblock Efficient decision-based black-box adversarial attacks on face recognition.
\newblock In \emph{CVPR}, pages 7714--7722, 2019.

\bibitem[Wang et~al.(2019)Wang, Liu, Jeon, Chu, and Matson]{wang2019end}
Yaqin Wang, Dongfang Liu, Hyewon Jeon, Zhiwei Chu, and Eric~T Matson.
\newblock End-to-end learning approach for autonomous driving: A convolutional neural network model.
\newblock In \emph{ICAART}, 2019.

\bibitem[Liu et~al.(2021{\natexlab{c}})Liu, Cui, Guo, Ding, Yang, and Chen]{liu2021visual}
Dongfang Liu, Yiming Cui, Xiaolei Guo, Wei Ding, Baijian Yang, and Yingjie Chen.
\newblock Visual localization for autonomous driving: Mapping the accurate location in the city maze.
\newblock In \emph{ICPR}, 2021{\natexlab{c}}.

\bibitem[Antoniak(1974)]{antoniak1974mixtures}
Charles~E Antoniak.
\newblock Mixtures of dirichlet processes with applications to bayesian nonparametric problems.
\newblock \emph{The annals of statistics}, pages 1152--1174, 1974.

\bibitem[Ferguson(1973)]{ferguson1973bayesian}
Thomas~S Ferguson.
\newblock A bayesian analysis of some nonparametric problems.
\newblock \emph{The annals of statistics}, pages 209--230, 1973.

\bibitem[Chang and Fisher~III(2013)]{chang2013parallel}
Jason Chang and John~W Fisher~III.
\newblock Parallel sampling of dp mixture models using sub-cluster splits.
\newblock In \emph{NeurIPS}, 2013.

\bibitem[Chen(2015)]{chen2015deep}
Gang Chen.
\newblock Deep learning with nonparametric clustering.
\newblock \emph{arXiv preprint arXiv:1501.03084}, 2015.

\bibitem[Dinari et~al.(2019)Dinari, Yu, Freifeld, and Fisher]{dinari2019distributed}
Or~Dinari, Angel Yu, Oren Freifeld, and John Fisher.
\newblock Distributed mcmc inference in dirichlet process mixture models using julia.
\newblock In \emph{2019 19th IEEE/ACM International Symposium on Cluster, Cloud and Grid Computing (CCGRID)}, pages 518--525, 2019.

\bibitem[Wang et~al.(2021{\natexlab{b}})Wang, Ni, Jing, Wang, Zhang, and Xing]{wang2021dnb}
Zeya Wang, Yang Ni, Baoyu Jing, Deqing Wang, Hao Zhang, and Eric Xing.
\newblock Dnb: A joint learning framework for deep bayesian nonparametric clustering.
\newblock \emph{TNNLS}, 33\penalty0 (12):\penalty0 7610--7620, 2021{\natexlab{b}}.

\bibitem[Blei and Jordan(2006)]{blei2006variational}
David~M Blei and Michael~I Jordan.
\newblock Variational inference for dirichlet process mixtures.
\newblock \emph{Bayesian analysis}, 2006.

\bibitem[Hoffman et~al.(2013)Hoffman, Blei, Wang, and Paisley]{hoffman2013stochastic}
Matthew~D Hoffman, David~M Blei, Chong Wang, and John Paisley.
\newblock Stochastic variational inference.
\newblock \emph{JMLR}, 2013.

\bibitem[Hughes and Sudderth(2013)]{hughes2013memoized}
Michael~C Hughes and Erik Sudderth.
\newblock Memoized online variational inference for dirichlet process mixture models.
\newblock In \emph{NeurIPS}, 2013.

\bibitem[Huynh et~al.(2016)Huynh, Phung, and Venkatesh]{huynh2016streaming}
Viet Huynh, Dinh Phung, and Svetha Venkatesh.
\newblock Streaming variational inference for dirichlet process mixtures.
\newblock In \emph{ACML}, pages 237--252, 2016.

\bibitem[Kurihara et~al.(2006)Kurihara, Welling, and Vlassis]{kurihara2006accelerated}
Kenichi Kurihara, Max Welling, and Nikos Vlassis.
\newblock Accelerated variational dirichlet process mixtures.
\newblock In \emph{NeurIPS}, 2006.

\bibitem[Ester et~al.(1996)Ester, Kriegel, Sander, Xu, et~al.]{ester1996density}
Martin Ester, Hans-Peter Kriegel, J{\"o}rg Sander, Xiaowei Xu, et~al.
\newblock A density-based algorithm for discovering clusters in large spatial databases with noise.
\newblock In \emph{KDD}, pages 226--231, 1996.

\bibitem[Shah and Koltun(2018)]{shah2018deep}
Sohil~Atul Shah and Vladlen Koltun.
\newblock Deep continuous clustering.
\newblock \emph{arXiv preprint arXiv:1803.01449}, 2018.

\bibitem[Tapaswi et~al.(2019)Tapaswi, Law, and Fidler]{tapaswi2019video}
Makarand Tapaswi, Marc~T Law, and Sanja Fidler.
\newblock Video face clustering with unknown number of clusters.
\newblock In \emph{ICCV}, pages 5027--5036, 2019.

\bibitem[Pakman et~al.(2020)Pakman, Wang, Mitelut, Lee, and Paninski]{pakman2020neural}
Ari Pakman, Yueqi Wang, Catalin Mitelut, JinHyung Lee, and Liam Paninski.
\newblock Neural clustering processes.
\newblock In \emph{ICML}, pages 7455--7465, 2020.

\bibitem[Ronen et~al.(2022)Ronen, Finder, and Freifeld]{ronen2022deepdpm}
Meitar Ronen, Shahaf~E Finder, and Oren Freifeld.
\newblock Deepdpm: Deep clustering with an unknown number of clusters.
\newblock In \emph{CVPR}, 2022.

\end{thebibliography}
}

\clearpage
\maketitlesupplementary

\renewcommand{\thesection}{\Alph{section}}
\setcounter{section}{0}
% \section*{Appendix}
For a better understanding of the main paper, we provide additional details in this supplementary material, which is organized as follows:
\begin{itemize}
\item \S\ref{sec:supp_a} provides the pseudo code of FEC.
\item \S\ref{sec:supp_b} introduces more experimental details. 
\item \S\ref{sec:supp_c} offers more results and discussions about the modeled representatives.
\item \S\ref{sec:supp_d} discusses our limitations, societal impact, and directions of future work.
\end{itemize}

\section{Pseudo Code} 
\label{sec:supp_a}
To facilitate a comprehensive understanding of FEC, we provide pseudo code for our feature encoding and feature pooling in Algorithm~\ref{alg:code}.

\section{More Experimental Detail} 
\label{sec:supp_b}

\noindent\textbf{Image Classification.}
In this task, several widely-used data augmentations are adopted to better train the model, including random horizontal flipping, random pixel erase~\citep{zhong2020random}, MixUp~\citep{zhang2017mixup}, CutMix~\citep{yun2019cutmix}, and label smoothing~\citep{szegedy2016rethinking}. We employ an AdamW~\citep{loshchilov2017decoupled} optimizer using a cosine decay learning rate scheduler and 5 epochs of warm-up. The momentum and weight decay are set to 0.9 and 0.05, respectively. A batch size of 1024 and an initial learning rate of 0.001 are used. We also use exponential moving average~\citep{polyak1992acceleration} to enhance the training. Throughput (image/s), or FPS, is measured using the same script~\citep{liu2021swin,touvron2021training} on a single V100 GPU using a batch size of 256. The reported values are averaged by 100 iterations after 20 warm iterations. We use the same codebase and tricks (\eg, multi-head computing) as in~\citep{ma2023image}. In addition, we use almost the same hyperparameters and architectures as in~\citep{ma2023image} for fair comparison.

\noindent\textbf{Downstream Tasks.}
During training, backbones are initialized with weights pre-trained on ImageNet~\citep{ImageNet}, while the other parts are initialized randomly.

\section{Modeled Representative} 
\label{sec:supp_c}

\begin{figure}[t]
    \renewcommand\thefigure{S1}
    \centering
   \includegraphics[width=0.8\linewidth]{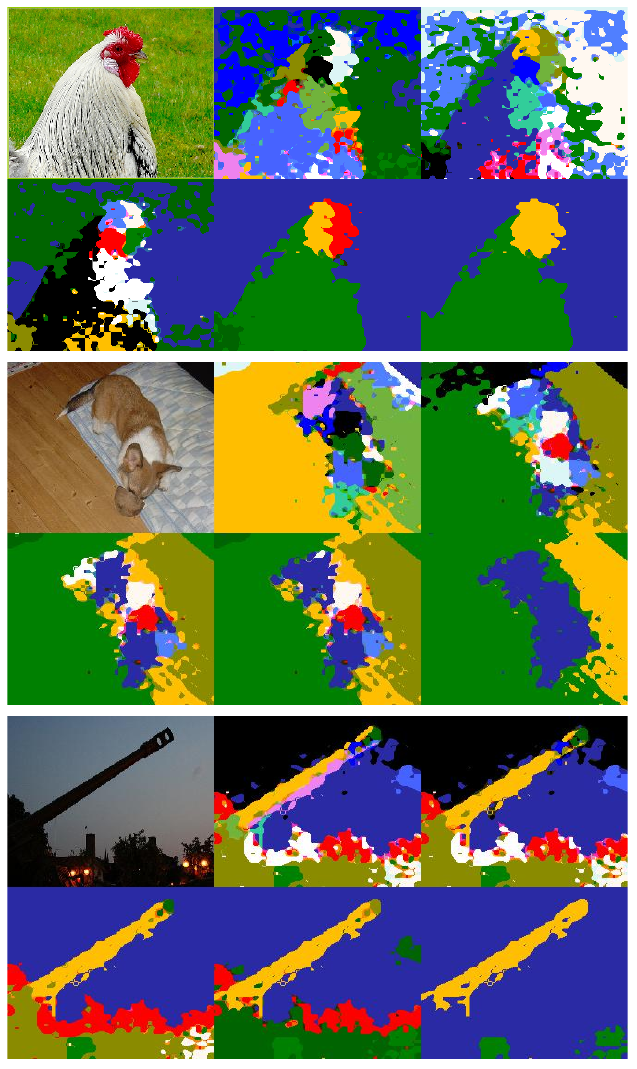}
   \caption{Inspection of the modeled representatives (\S\ref{sec:supp_c}) on ImageNet-1K~\citep{ImageNet} \texttt{val}. }
\label{fig:rep_supp1}
\end{figure}

In the ``Study of \emph{Ad-hoc} Interpretability'' section of the main paper, it is highlighted that FEC's final cluster assignments display consistent semantic representations. These representations frequently correlate with distinct objects or their components and demonstrate a close alignment with human perception. Here we visualize more results of cluster assignments in Fig.~\ref{fig:rep_supp1} to clarify the FEC's principles. Similar conclusions can be drawn from Fig.~\ref{fig:rep_supp1}, which confirms again the \emph{ad-hoc} interpretability and effectiveness of FEC.

\begin{algorithm*}[t]
    \renewcommand\thealgorithm{S1}
    \caption{Pseudo code of FEC in a PyTorch-like style.}
    \label{alg:code}
    \definecolor{codeblue}{rgb}{0.25,0.5,0.5}
	\lstset{
		backgroundcolor=\color{white},
		basicstyle=\fontsize{7.8pt}{7.8pt}\ttfamily\selectfont,
		columns=fullflexible,
		breaklines=true,
		captionpos=b,
		escapeinside={(:}{:)},
		commentstyle=\fontsize{7.8pt}{7.8pt}\color{codeblue},
		keywordstyle=\fontsize{7.8pt}{7.8pt},
		%  frame=tb,
	}
   \begin{lstlisting}[language=python]
   # feat_i: input feature (N x C), where N = W x H
   
   # C: number of channels
   # N: resolution of input feature
   # O: number of cluster centers. In pooling, O = N/4. In encoding, O is a hyperparameter (O < N).
   # M: similarity matrix (Eq.5)
   # A: cluster assigment matrix
   # R: representatives
   # sig: sigmoid function
   # alpha and beta: learnable parameters

   def model_representatives(feat_i)
       # center initialization (Eq.4)
       feat_k = conv_k(feat_i)  # (N x C')
       feat_v = conv_v(feat_i)  # (N x C')
       feat_c_k = ada_pool(feat_k)  # (O x C')
       feat_c_v = ada_pool(feat_v)  # (O x C')

       # compute similarities and cluster assigments (Eq.5)
       M = cosine_sim(feat_k, feat_c_k)  # (N x O)
       A = torch.argmax(M, dim=1)  # (N x O)

       # aggragate the feature of representatives (Eq.6)
       R = aggragate_feature(feat_v, feat_c_v, A)  # (O x C')

       return R, M

   def pooling(feat_i)
       R, _ = model_representatives(feat_i)
       res_conn = ResConn(feat_i)  # residual connection (Eq.9)
       
       return R + res_conn

   def encoding(feat_i)
       R, M = model_representatives(feat_i)

       # feature dispatching (Eq.7)
       refined_M = sig(alpha * M + beta).permute(1,0)  # (O x N)
       feat_d = ( R.unsqueeze(dim=1) * refined_M.unsqueeze(dim=-1) ).sum(dim=0)  # (N x C')
       feat_d = MLP(feat_d)  # (N x C)
       out = feat_i + feat_d  # residual connection
       
       return out
   \end{lstlisting}
\end{algorithm*}

\section{Discussion} 
\label{sec:supp_d}

\noindent\textbf{Limitation Analysis.}
One limitation of our approach is the adoption of a straightforward clustering mechanism, primarily aimed at ensuring computational efficiency. While this design choice contributes to faster processing times, it may inadvertently lead to sub-optimal performance in certain scenarios. Additionally, akin to many parametric clustering algorithms~\citep{mitchell1999machine,xu2005survey,van2020scan}, our method requires the manual definition of the number of clusters to keep the same resolution with previous works~\citep{he2016deep,liu2021swin,touvron2022resmlp,yu2022metaformer}. This aspect introduces a degree of subjectivity and potential bias, as the predetermined cluster count may not align perfectly with the intrinsic structure of specific images, particularly in dealing with datasets where the optimal number of clusters is not known a priori or varies significantly.

\noindent\textbf{Societal Impact.}
This work provides a clustering perspective for transparent, \emph{ad-hoc} interpretable feature extraction, and accordingly introduces a novel visual backbone which reformulates the entire process of feature extraction as representative selection. On positive side, the approach advances network interpretability and is valuable in safety-sensitive applications, \eg, medical image analysis~\citep{duncan2000medical}, face recognition~\citep{yang2020design,dong2019efficient}, and autonomous driving~\citep{wang2019end,liu2021visual}. For potential negative social impact, the erroneous recognition may cause inaccurate decision or planning of systems based on the results. In addition, the potential bias inherent in the training data may be exploited for malicious purposes.

\noindent\textbf{Future Work.}
This work also comes with new challenges, certainly worth further exploration:
\begin{itemize}
    \item \textbf{Incorporating Advanced Clustering Algorithms}. In future developments, we aim to augment the FEC framework by incorporating advanced clustering algorithms. Our current model prioritizes computational efficiency with a straightforward clustering mechanism, but we recognize opportunities for enhancing performance and accuracy. Upcoming versions will investigate sophisticated algorithms adept at managing complex data structures and distributions, potentially increasing the granularity and precision of feature extraction for more refined and accurate visual representations. An intriguing avenue is transitioning from parametric clustering, which presupposes a fixed number of clusters, to nonparametric clustering, where the number of clusters is undetermined. There are numerous techniques for nonparametric clustering, including Bayesian nonparametric (BNP) mixture models (exemplified by the Dirichlet Process Mixture (DPM) model~\citep{antoniak1974mixtures,ferguson1973bayesian}), DPM sampler~\citep{chang2013parallel,chen2015deep,dinari2019distributed,wang2021dnb}, variational DPM inference~\citep{blei2006variational,hoffman2013stochastic,hughes2013memoized,huynh2016streaming,kurihara2006accelerated}, density-based approach~\citep{ester1996density}, nearest-neighbor graph~\citep{shah2018deep}, supervised approach~\citep{tapaswi2019video,pakman2020neural}, dynamic network architecture~\citep{ronen2022deepdpm}. We have explored a very recent work, \ie, DeepDPM~\citep{ronen2022deepdpm}. However, after running their code, we find that DeepDPM is notably complex and require substantial computational time. Moving forward, our focus is on identifying better trade-offs between complexity, computational efficiency, and performance.
    \item \textbf{Combination with Set-prediction Architectures}. The recent emergence of set-prediction architectures, such as DETR~\citep{carion2020end}, presents a significant opportunity to utilize the representatives modeled by FEC more effectively. Unlike traditional methods that rely on hand-crafted components like non-maximum suppression for post-processing and pre-defined anchors for label assignments, these approaches simplify the pipeline by allowing for end-to-end training and inference. This reduces the need for many of the specialized components typically used in object detection systems and provides an ideal framework for utilizing the representatives extracted by FEC. For example, the modeled representatives can be applied as a metric for distance measurement, aiding in the stabilization of bipartite matching. This integration effectively infuses the concept of ``instances'' (or representatives) into the feature extraction process, which stands as the primary motivation behind this work.
\end{itemize}

\end{document}